\documentclass{article}

\PassOptionsToPackage{numbers,compress}{natbib}

\usepackage[preprint]{neurips_2026}
\usepackage[utf8]{inputenc}
\usepackage[T1]{fontenc}
\usepackage{nicefrac}
\usepackage{microtype}
\usepackage{url}

\usepackage{eccvabbrv}

\usepackage{graphicx}
\usepackage{wrapfig}
\usepackage{placeins}
\usepackage{booktabs}
\usepackage{multirow} % For spanning rows
\usepackage{graphicx} % For \resizebox
\usepackage{amsmath}
\usepackage{algorithm}
\usepackage{algpseudocode} % from algorithmicx
\usepackage[dvipsnames]{xcolor}        % for \textcolor comments
\usepackage{caption}                    % for \captionof inside wrapfigure
\newcommand{\myparagraph}[1]{\smallskip\noindent\textbf{#1}}

\usepackage[accsupp]{axessibility}  % Improves PDF readability for those with disabilities.

\usepackage{hyperref}
\usepackage[capitalize]{cleveref}

\usepackage{orcidlink}

\title{Topo-R1: Detecting Topological Anomalies via Vision-Language Models}

\author{%
  Meilong Xu\textsuperscript{1} \quad
  Qingqiao Hu\textsuperscript{1} \quad
  Xiaoling Hu\textsuperscript{2} \quad
  Shahira Abousamra\textsuperscript{3} \quad
  Xin Yu\textsuperscript{4} \\
  \textbf{Weimin Lyu\textsuperscript{1}} \quad
  \textbf{Kehan Qi\textsuperscript{1}} \quad
  \textbf{Dimitris Samaras\textsuperscript{1}} \quad
  \textbf{Chao Chen\textsuperscript{1}} \\[4pt]
  \textsuperscript{1}Stony Brook University \quad
  \textsuperscript{2}Massachusetts General Hospital and Harvard Medical School \\
  \textsuperscript{3}Stanford University \quad
  \textsuperscript{4}Penn State University
}

\begin{document}

\maketitle
\renewcommand{\thefootnote}{\fnsymbol{footnote}}
\footnotetext{Email: Meilong Xu (meixu@cs.stonybrook.edu).}
\renewcommand{\thefootnote}{\arabic{footnote}}

\begin{abstract}
Topology plays a critical role in tubular structures such as blood vessels, nerve fibers, and road networks, where connectivity and loop structure directly govern downstream functional analysis. Vision-Language Models (VLMs) are promising candidates for understanding such structures, given their visual reasoning and grounding capabilities. To probe their topological perception, we systematically evaluate leading closed-source and state-of-the-art open-source VLMs on localizing and classifying four canonical topological anomalies (broken or spurious connections, missing or extra branches) in tubular-network segmentation masks. We find that they perform nearly at random, indicating that topology-aware perception is largely absent from current general-purpose VLMs. As no existing resource pairs segmentation masks with localized anomaly annotations, we build an automated, multi-domain data-curation pipeline that synthesizes diverse topological perturbations with verifiable Betti-number annotations across graduated difficulty levels, yielding the first systematic benchmark with a large-scale training set and held-out in-distribution and out-of-distribution test suites. Building on this benchmark, we introduce \textbf{Topo-R1}, centered on a topology-aware composite reward that jointly scores localization, classification, and skeleton-level structural fidelity. We use supervised fine-tuning as a cold start to bootstrap schema-compliant outputs, then optimize the policy against this reward via Group Relative Policy Optimization (GRPO), steering predictions toward topologically meaningful structures rather than superficial pixel overlap. Extensive experiments show that \textbf{Topo-R1} substantially outperforms general-purpose VLMs and matches or exceeds supervised baselines across in-distribution, out-of-distribution, and real-segmentation-output protocols, establishing a strong foundation for VLM-based topological understanding of structured visual data.
% \keywords{Topological Data Analysis, Anomaly Detection, Vision Language Model}
\end{abstract}

\section{Introduction}
\label{sec:intro}

\begin{figure*}[t]
    \centering
    \includegraphics[width=\textwidth]{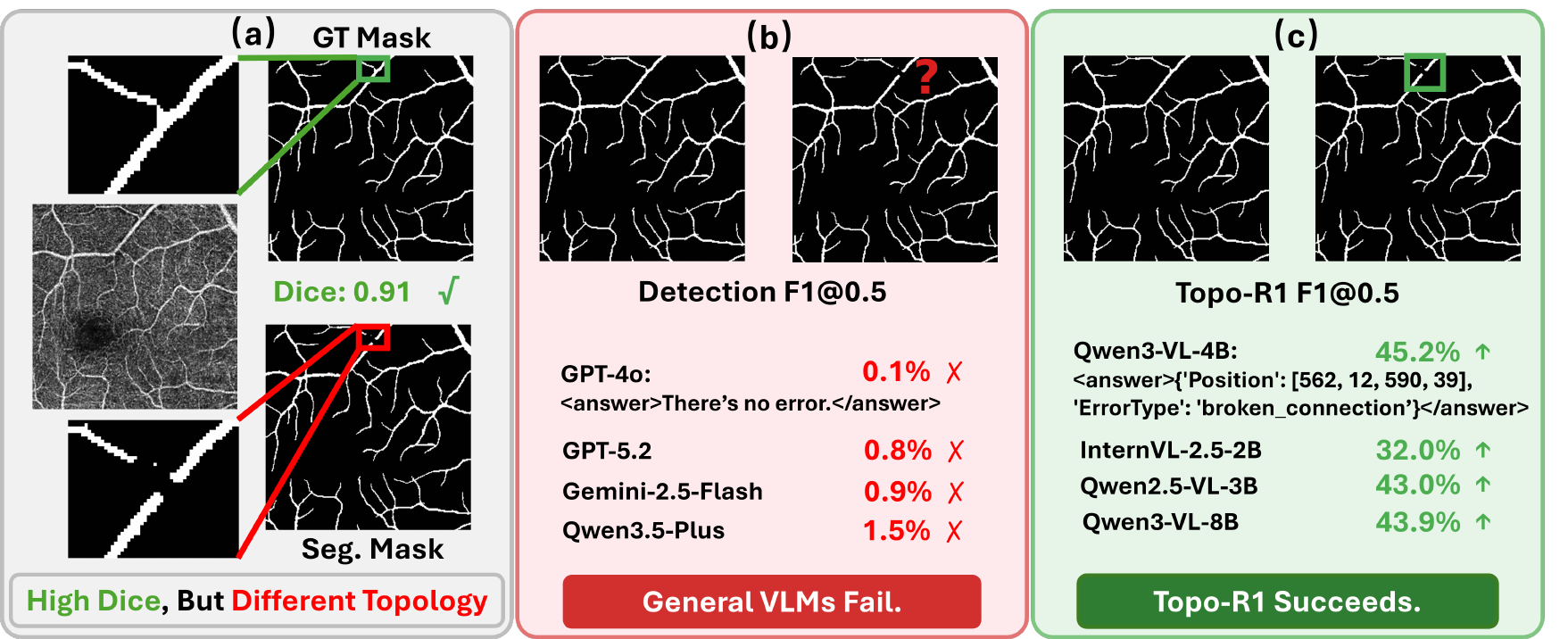}
    \caption{\textbf{Intuition of the framework.} (a)~A segmentation mask can achieve a high Dice score (0.91) yet contain critical topological anomalies, such as broken or spurious connections, that are visible only upon close structural inspection. (b)~State-of-the-art VLMs, including GPT-5.2 and Gemini-2.5-Flash, fail to detect these topological anomalies (near-zero Detection F1@0.5). (c)~Topo-R1 successfully detects and classifies the topological anomaly with structured, typed bounding-box outputs.}
    \label{fig:teaser}
\end{figure*}

Tubular structures, such as blood vessels, nerve fibers, and road networks, underlie biological systems and human infrastructure~\cite{mosinska2018beyond, hu2019topology, shit2021cldice, khandouzi2022retinal, liu2024deep}. While morphological attributes such as thickness and branching density contribute to the analysis of these structures, topology plays a uniquely critical role: connectivity and loop structure directly govern blood flow, neural signaling, and route planning. A single missing pixel can sever a vessel and invalidate a hemodynamic simulation; a spurious bridge between two road segments can corrupt an entire navigation graph. A model that can perceive such topological patterns therefore unlocks a broad range of downstream applications, from quality assurance of existing segmentation pipelines to self-training and active learning across heterogeneous imaging modalities.

Yet topology-aware perception remains a long-standing open challenge. Existing topology-preserving methods~\cite{hu2019topology, shit2021cldice, stucki2023topologically, clough2020topological, hu2022structure} fundamentally rely on supervised training with pixel-level ground-truth annotations, and even recent semi-supervised extensions~\cite{xu2024semi, xu2025match} still require a labeled subset within the same domain. Producing such annotations requires specialized domain expertise and is extremely time-consuming, and the large domain gap across application areas (e.g., retinal vasculature, road networks, neural circuits) means that annotations from one domain rarely transfer to another. This raises the question of whether a more general model can directly learn topological patterns of tubular structures, without being tied to a single domain or relying on dense pixel-level supervision.

\begin{wrapfigure}{r}{0.42\textwidth}
    \centering
    \includegraphics[width=\linewidth]{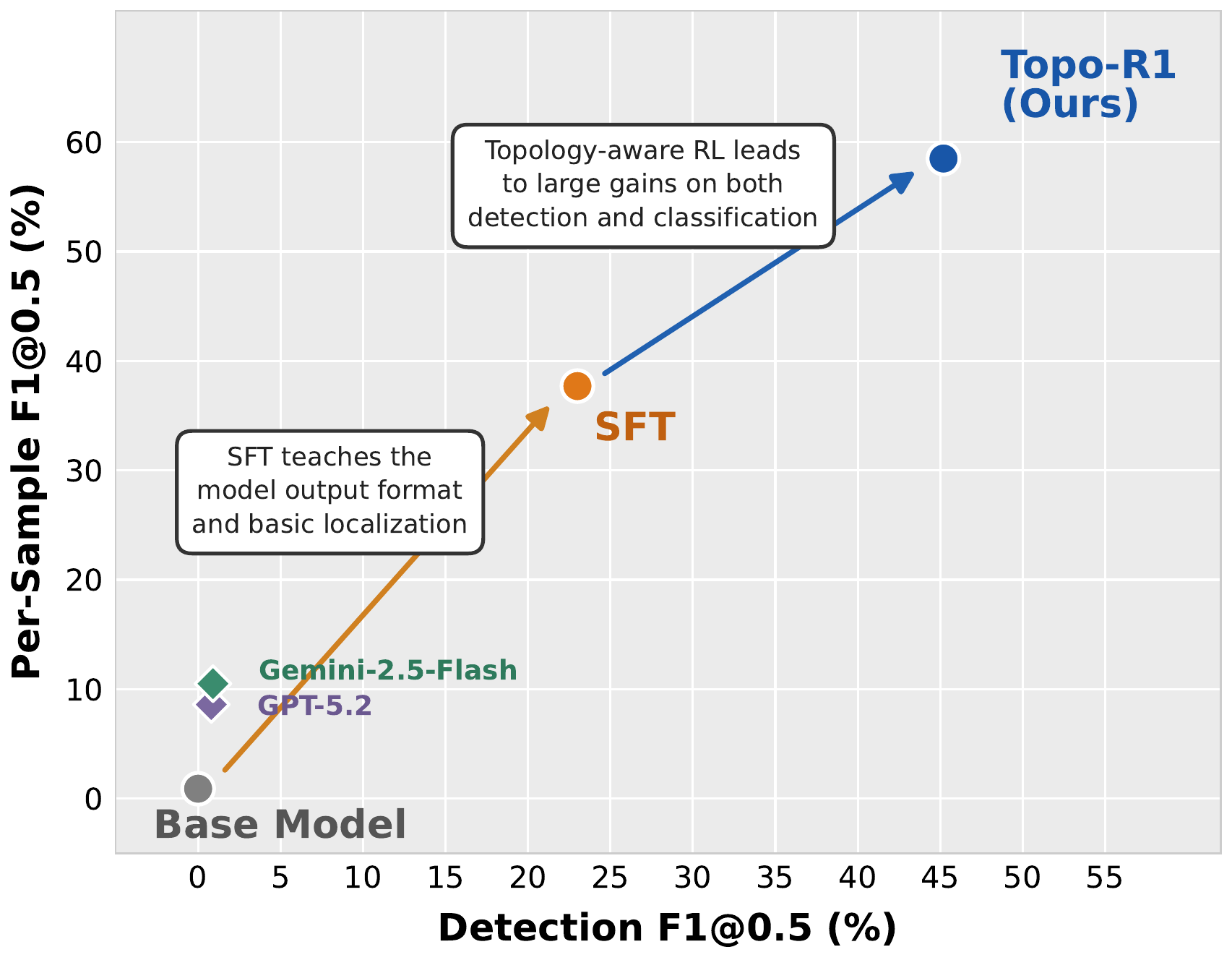}
    \caption{The performance of state-of-the-art VLMs and our Topo-R1 on the topological anomaly detection task.}
    \label{fig:performance}
\end{wrapfigure}
Vision-Language Models (VLMs) such as GPT-4o~\cite{hurst2024gpt}, Gemini~\cite{team2023gemini}, LLaVA~\cite{liu2023visual, liu2024improved}, Qwen-VL~\cite{bai2023qwenvl, bai2025qwen2, wang2024qwen2, qwen3technicalreport}, InternVL~\cite{chen2024internvl}, and Molmo~\cite{deitke2025molmo} are powerful candidates for this challenge, given their strong capabilities across visual question answering, grounding, and spatial reasoning~\cite{li2023llavamed, zhang2024biomedclip, moor2023med}. To probe whether they already possess topological pattern understanding, we cast topology-aware perception as the task of localizing and classifying \textbf{topological anomalies}, defined as localized regions where the topology of a predicted segmentation mask is inconsistent with the underlying tubular-network topology depicted by the input image, with the image itself serving as the contextual reference (formal definition in \cref{subsec:data}). As shown in~\cref{fig:teaser}, our systematic evaluation reveals that both leading closed-source and state-of-the-art open-source VLMs perform almost at random on this probe: even with carefully designed prompts and in-context examples, they consistently fail to identify \textbf{localized yet important connectivity changes} such as a few missing pixels that sever a vessel or a thin bridge that wrongly merges two road segments, indicating that topology-aware perception is largely absent from current general-purpose VLMs.

This gap stems from limitations in both training data and the inherent structure of the problem.
First, public-domain VLMs are typically pretrained on data emphasizing medium- and large-scale visual patterns, with virtually no training signal for fine-grained topological perception~\cite{tong2024cambrian, karamcheti2024prismatic, chen2024far}.
More fundamentally, topological patterns in tubular networks are extremely sparse and localized: a single missing pixel among thousands of correctly segmented ones can break a critical connection. Recognizing such patterns requires the model to efficiently search a vast, mostly correct structure and identify the few points where connectivity changes, a needle-in-a-haystack problem that standard VLMs have no mechanism to solve. Unlike conventional object detection, where salient visual cues guide attention, topological cues are often visually subtle and can only be recognized by tracing long-range connectivity across the entire network, demanding a combination of global structural reasoning and fine-grained local perception that current architectures lack.

To endow VLMs with topology-aware perception, we observe that no existing resource pairs segmentation masks with the localized topological-anomaly annotations needed for both training and evaluation. We therefore develop an automated, multi-domain data-curation pipeline that injects four types of controlled topological perturbations into clean masks, with automated Betti-number verification, across graduated difficulty levels. The pipeline yields the first systematic benchmark for VLM-based topological perception of tubular structures, with a large-scale training set and a held-out test suite covering both in-distribution and out-of-distribution settings. Built on top of this benchmark, we introduce \textbf{Topo-R1}, a vision-language framework that learns topological patterns through localizing and classifying topological anomalies in segmentation masks. \textbf{Topo-R1} adopts a two-stage training paradigm of supervised fine-tuning followed by reinforcement learning with Group Relative Policy Optimization (GRPO)~\cite{shao2024deepseekmath}, and as shown in~\cref{fig:performance}, substantially outperforms both general-purpose VLMs and supervised baselines across the benchmark; we additionally stress-test on outputs of real segmentation models and observe that the gains transfer to errors actually produced in deployment.

In summary, our contributions are threefold:

\begin{enumerate}
    \item We construct the first systematic benchmark for VLM-based topological perception of tubular structures, generated through an automated, multi-domain data-curation pipeline and comprising a large-scale training set with a held-out test suite that covers both in-distribution and out-of-distribution settings; both the pipeline and benchmark will be publicly released.
    \item Through the systematic evaluation enabled by this benchmark, we identify that current general-purpose VLMs, including both leading closed-source and state-of-the-art open-source models, lack topology-aware perception. We accordingly introduce \textbf{Topo-R1}, the first vision-language framework that equips VLMs with this capability, opening a new direction of using VLMs for fine-grained, structure-aware visual analysis of tubular networks.
    \item Extensive experiments across multiple VLM backbones and imaging domains demonstrate that \textbf{Topo-R1} substantially outperforms both general-purpose VLMs and supervised baselines, generalizes to out-of-distribution domains, and remains robust when stress-tested on outputs from real segmentation models.
\end{enumerate}

\section{Related Work}
\label{sec:related}
\myparagraph{Topology-Driven Deep Image Analysis.}
Tubular structure segmentation has evolved from hand-crafted filters~\cite{frangi1998multiscale} and encoder-decoder architectures~\cite{ronneberger2015u,zhou2020unetplus,isensee2021nnu,mou2021cs2net} to topology-aware losses such as persistent-homology penalties~\cite{bentaieb2016topology,hu2019topology,clough2020topological,xu2024semi,xu2025match}, clDice~\cite{shit2021cldice}, Skeleton Recall~\cite{kirchhoff2024skeleton}, homotopy warping~\cite{hu2022structure}, Betti matching~\cite{stucki2023topologically,stucki2024efficient,lux2024topograph}, and Euler-characteristic optimisation~\cite{li2025topology}; Decroocq~et~al.~\cite{decroocq2025benchmarking} benchmark these metrics comprehensively. All such methods shape the training loss to suppress topological violations but do not reason about where each violation occurs or what type it is. For \emph{post-hoc} assessment, Li~et~al.~\cite{li2023robust,li2024universal} detect violations via Euler Characteristic maps, yet produce only pixel-level heatmaps without structured, typed detections and remain coupled to specific CNN backbones. Our work instead formulates topological anomaly detection as a structured visual reasoning task with typed bounding-box output.

\myparagraph{VLMs for Visual Understanding and Inspection.}
Vision-Language Models (VLMs) align visual encoders with LLMs via contrastive pretraining~\cite{radford2021learning}, cross-attention~\cite{alayrac2022flamingo}, bridging modules~\cite{li2023blip2,dai2023instructblip}, and visual instruction tuning~\cite{liu2023visual,liu2024improved,zhu2023minigpt4}, yielding systems such as Qwen-VL~\cite{bai2023qwenvl,bai2025qwen2}, InternVL~\cite{chen2024internvl}, CogVLM~\cite{wang2023cogvlm}, GPT-4V~\cite{openai2023gpt4v}, and Gemini~\cite{team2023gemini}. Grounding-oriented models~\cite{peng2023kosmos2,chen2023shikra,you2024ferret,liu2023grounding} add region-level perception, and domain-specific adaptations address biomedical VQA~\cite{li2023llavamed} and reasoning-based segmentation~\cite{lai2024lisa}. A parallel line applies VLMs as visual quality inspectors: AnomalyGPT~\cite{gu2024anomalygpt} and MMAD~\cite{jiang2024mmad} target industrial anomaly detection; Anomaly-OneVision~\cite{xu2025towards} and AnomalyR1~\cite{chao2025anomalyr1} extend this to zero-shot and RL-based settings; and ConnectomeBench~\cite{brown2025connectomebench} benchmarks LLMs on neuron segmentation proofreading but provides only a 3D benchmark, not a trainable detector for 2D tubular topological anomalies. Our work addresses this gap by introducing a VLM that detects and classifies topological anomalies using structured bounding-box outputs.

\myparagraph{Reinforcement Fine-Tuning in Visual Context.}
RL-based alignment has evolved from RLHF with PPO~\cite{christiano2017deep,ouyang2022training,schulman2017proximal} and preference methods~\cite{rafailov2023direct,azar2024general,ethayarajh2024kto} to GRPO~\cite{shao2024deepseekmath}, which enables compact visual reasoning models~\cite{huang2025vision,yang2025r1,zhang2025r1,xia2025visionary,tan2025reason,feng2025video} to rival larger supervised counterparts, with stability improvements from Hint-GRPO~\cite{huang2025boosting} and VL-Rethinker~\cite{wang2025vl}. For fine-grained perception, Visual-RFT~\cite{liu2025visual} pioneered IoU-based rewards; Seg-Zero~\cite{liu2025seg}, VisionReasoner~\cite{liu2026visionreasonerq}, and LENS~\cite{zhu2025lens} extend RL to segmentation; VLM-R1~\cite{shen2025vlm} and Perception-R1~\cite{yu2025perception} highlight reward design as the pivotal factor; and grounding emerges from answer-correctness rewards alone~\cite{sarch2025grounded,fan2025grit,shen2025satori,batra2025spatialthinker}. Medical adaptations~\cite{pan2025medvlm,lai2025med,su2025gmai,xu2025medground} apply GRPO to visual reasoning and grounding, while preference-based alternatives~\cite{sun2023aligning,yu2024rlhf,yu2024rlaif,wang2024mdpo,li2024silkie} target hallucination reduction. Despite these advances, no prior work applies RL fine-tuning to topological quality assessment. We bridge this gap with a composite GRPO reward that incorporates a type-aware centerline Dice component and a domain-specific formulation unexplored in prior visual RL.

\section{Methodology}
\label{sec:method}
In this section, we will introduce our \textbf{Topo-R1} framework in detail.

\subsection{Preliminaries}
\label{subsec:preliminaries}
\myparagraph{Topological Data Analysis.}
Topological data analysis (TDA)~\cite{edelsbrunner2010computational} characterizes the shape and connectivity of data through \emph{Betti numbers} $\beta_k=\mathrm{rank}\,H_k(X)$, which count independent $k$-dimensional holes. In 2D tubular-network segmentation, $\beta_0$ reflects the number of disjoint fragments and $\beta_1$ the number of closed loops, making Betti-number changes a natural proxy for topological correctness~\cite{hu2019topology, stucki2023topologically}.

\myparagraph{Group Relative Policy Optimization.}
Supervised fine-tuning (SFT) optimizes the next-token likelihood $\mathcal{L}_{\text{SFT}} = -\sum_{t} \log \pi_\theta(y_t \mid x, y_{<t})$, which does not directly capture downstream goals such as precise localization and accurate type classification. We therefore refine the policy with Group Relative Policy Optimization (GRPO)~\cite{shao2024deepseekmath}, which forgoes a separate critic network by estimating advantages from within-group reward statistics. For each query $q$, GRPO samples $G$ candidate outputs from $\pi_\theta$, scores each by the composite reward of~\cref{subsec:reward} to obtain rewards $\{R_1, \dots, R_G\}$, and standardizes within the group to form per-sample advantages $\hat{A}_i = (R_i - \mu_G)/(\sigma_G + \epsilon_{\text{std}})$, where $\mu_G,\sigma_G$ are the group mean and standard deviation. The policy is then updated by maximizing a clipped PPO-style surrogate~\cite{schulman2017proximal} with a KL-divergence penalty toward a reference policy $\pi_{\mathrm{ref}}$ initialized from the SFT checkpoint; full equations for the GRPO objective, the per-sample clipped surrogate, and the importance-sampling ratio are provided in Appendix~\ref{sec:supp_impl}.

\subsection{Automated Data Curation Pipeline}
\label{subsec:data}

\myparagraph{Task Formulation and Error Taxonomy.}
Given an input image $I \in \mathbb{R}^{H \times W \times 3}$ and a binary segmentation mask $M \in \{0,1\}^{H \times W}$, topological anomaly detection produces detections $\mathcal{E} = \{(b_i, t_i)\}_{i=1}^{N}$, where $b_i$ is the bounding box of the $i$-th \textbf{potential error} and $t_i \in \mathcal{T}$ its type ($\mathcal{E}\!=\!\emptyset$ if none). Each detection flags a region where the topology of $M$ is inconsistent with that of the tubular network depicted in $I$ (the \textbf{contextual reference}). The model receives the pair $(I, M)$ at both training and test time; only during training is the clean ground-truth mask additionally available, against which labels are derived by symmetric topological differencing. The taxonomy $\mathcal{T}$ has four categories, exhaustive over local perturbations and verifiable via Betti-number changes:
\begin{itemize}
    \item \textbf{Broken connection}: a gap severing a continuous segment ($\beta_0\!\uparrow$ or $\beta_1\!\downarrow$).
    \item \textbf{Spurious connection}: a bridge merging distinct segments ($\beta_0\!\downarrow$ or $\beta_1\!\uparrow$).
    \item \textbf{Missing branch}: a terminal branch present in the ground truth but absent from $M$, reducing branching.
    \item \textbf{Extra branch}: a false branch present in $M$ but absent from the ground truth, inflating branching.
\end{itemize}

\begin{figure*}[t]
    \centering
    \includegraphics[width=\textwidth]{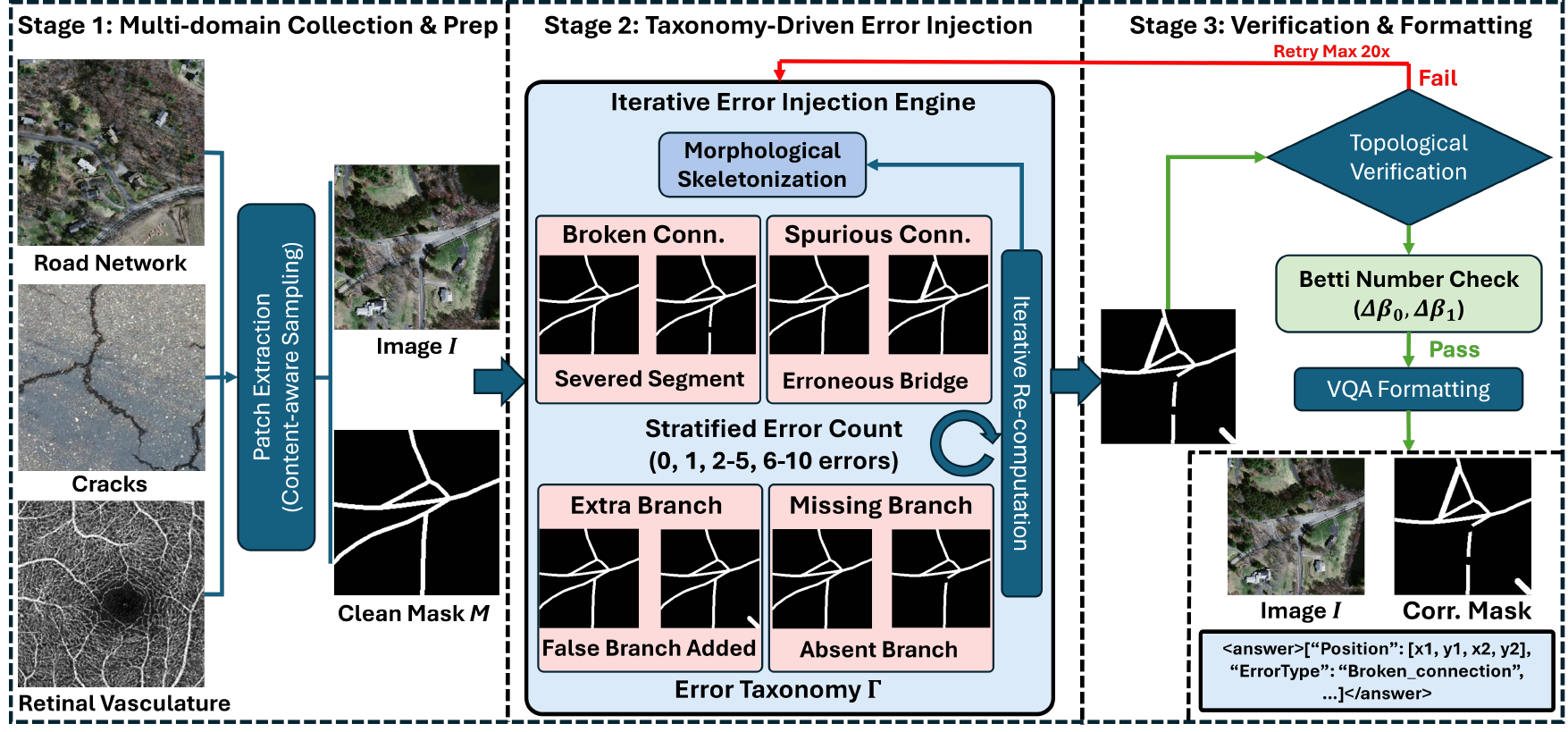}
    \caption{Overview of the automatic data curation pipeline.}
    \label{fig:data_curation}
\end{figure*}

\myparagraph{Curation Pipeline.}
Manual annotation of these anomalies is expensive at scale, so we develop a fully automated pipeline (\cref{fig:data_curation}). We aggregate data from multiple public sources, \eg, aerial roads, surface cracks, retinal vasculature, leaf-veins, and retinal fundus (per-source breakdown in Appendix~\ref{sec:supp_split} and~\cref{tab:per_dataset}), extract $256\!\times\!256$ patches via content-aware sampling, and inject controlled anomalies into clean masks under a complexity-stratified curriculum. Crucially, \textbf{every accepted injection is verified by Betti-number checking to alter $\beta_0$ or $\beta_1$}, guaranteeing that each labeled anomaly corresponds to a genuine topological change; failed operator applications are rejected and retried. Each sample pairs the patch with its (corrupted or clean) mask and a detection prompt; the ground-truth answer is a list of typed bounding-box dictionaries in \texttt{<answer>} tags. Full strata percentages, type-balancing sampler, injection operators, retry policy, and bounding-box formatting are deferred to Appendix~\ref{sec:supp_split}.

\subsection{Reward Design}
\label{subsec:reward}
The reward function shapes the policy's localization and classification behavior. We design a composite reward
\begin{equation*}
    R_{\text{total}} = w_{\text{fmt}}\,R_{\text{fmt}} + w_{\text{acc}}\,R_{\text{acc}} + w_{\text{topo}}\,R_{\text{topo}},
\end{equation*}
with weights $(w_{\text{fmt}}, w_{\text{acc}}, w_{\text{topo}}) = (0.10, 0.85, 0.05)$ summing to one. The format reward $R_{\text{fmt}}\!\in\!\{0,1\}$ is a binary gate verifying that the output parses as a valid list of $\{\texttt{Position},\texttt{ErrorType}\}$ dictionaries with labels in $\mathcal{T}$ and integer coordinates in $[0,1000]$; the remaining sub-rewards are computed only on schema-compliant outputs.

\myparagraph{Accuracy Reward.}
The accuracy reward $R_{\text{acc}} = R_{\text{det}} + R_{\text{loc}} + R_{\text{type}}$ aggregates three sub-objectives over predictions and ground-truth errors paired by \textbf{type-aware Hungarian matching}: for each type $t \in \mathcal{T}$ we solve a per-type linear assignment with IoU as the cost, accept matches with $\mathrm{IoU} \ge \tau_m = 0.1$, and treat unmatched predictions as false positives ($\mathrm{FP}$) and unmatched ground-truth errors as false negatives ($\mathrm{FN}$). Matching is per-type by design: a prediction with the wrong type can never be matched to a ground-truth error of the correct type, so type errors automatically inflate both $\mathrm{FP}$ and $\mathrm{FN}$, directly coupling classification correctness to detection. As corner cases, $R_{\text{acc}}=1$ if both sides are empty, $R_{\text{acc}}=0$ if exactly one side is empty, and $R_{\text{loc}}=R_{\text{topo}}=0$ when $|\mathcal{M}|=0$. The detection reward is a soft F1:
\begin{equation*}
    R_{\text{det}} = \frac{2\,\widetilde{\mathrm{TP}}}{2\,\widetilde{\mathrm{TP}}+|\mathrm{FP}|+|\mathrm{FN}|},\qquad
    \widetilde{\mathrm{TP}} = \sum_{(i,j)\in\mathcal{M}} \phi(\mathrm{IoU}_{ij}),
\end{equation*}
where $\phi$ smoothly maps IoU to a reward score via piecewise non-linear interpolation across tier thresholds $\{\tau_k\}=\{0.3,0.5,0.7,0.9\}$ with reward ceilings $\{r_k\}=\{0.25,0.55,0.80,1.0\}$ and exponent $\gamma=1.5$:
\begin{equation*}
    \phi(\mathrm{IoU}) = \begin{cases}
        r_1\,(\mathrm{IoU}/\tau_1)^{\gamma} & \mathrm{IoU}<\tau_1\\[2pt]
        r_{k-1}+(r_k-r_{k-1})\!\left(\dfrac{\mathrm{IoU}-\tau_{k-1}}{\tau_k-\tau_{k-1}}\right)^{\!\gamma} & \tau_{k-1}\le\mathrm{IoU}<\tau_k\\[2pt]
        r_K & \mathrm{IoU}\ge\tau_K.
    \end{cases}
\end{equation*}
\textbf{$R_{\text{det}}$ is structurally identical to the F1 score the model is evaluated on}, so the policy is trained on exactly the metric it is measured by; the smooth $\phi$ further provides dense gradient signal across tier transitions rather than the all-or-nothing reward of a hard IoU threshold. The localization reward $R_{\text{loc}} = \frac{1}{|\mathcal{M}|}\sum_{(i,j)\in\mathcal{M}}\phi(\mathrm{IoU}_{ij})$ averages the same $\phi(\mathrm{IoU})$ values that compose $\widetilde{\mathrm{TP}}$ in $R_{\text{det}}$, providing a per-match localization signal complementary to the precision/recall trade-off of $R_{\text{det}}$. The type bonus $R_{\text{type}} = |\mathcal{M}|/\max(N_{\text{gt}},N_{\text{pred}})$ rewards joint type-and-localization accuracy, since per-type matching can only produce $|\mathcal{M}|$ matches when both type and location agree.

\myparagraph{Topological Reward.}
IoU is a pixel-area measure that is blind to whether a region contains topologically meaningful content. We add a centerline-Dice (clDice)~\cite{shit2021cldice} reward computed only over Hungarian-matched pairs, so it inherits the type-awareness of the matching stage. For a matched pair $(i,j)$ with cropped masks $M_i^{\text{gt}}, M_j^{\text{corr}}$ and their skeletons $S_i^{\text{gt}}, S_j^{\text{corr}}$, the standard clDice is the harmonic mean of $\mathrm{Tprec}=|S_j^{\text{corr}}\cap M_i^{\text{gt}}|/|S_j^{\text{corr}}|$ and $\mathrm{Tsens}=|S_i^{\text{gt}}\cap M_j^{\text{corr}}|/|S_i^{\text{gt}}|$. Regions containing topological anomalies exhibit low clDice between corrupted and ground-truth masks, so the per-pair reward is $(1-\mathrm{clDice})$, modulated by a localization penalty $\mathrm{LocPen}(b)\!\in\![0,1]$ that linearly attenuates the reward once the box exceeds $\tau_{\text{size}}\!=\!0.3$ of the image area (slope $\lambda\!=\!0.8$; closed form in Appendix~\ref{sec:supp_impl}), suppressing oversized boxes which trivially capture topological changes. The aggregate is
\begin{equation*}
    R_{\text{topo}} = \frac{1}{|\mathcal{M}|}\sum_{(i,j)\in\mathcal{M}}\big(1-\mathrm{clDice}(M_i^{\text{gt}},M_j^{\text{corr}})\big)\cdot\mathrm{LocPen}(b_j^{\text{pred}}).
\end{equation*}
\textbf{Type conditioning prevents the policy from gaming the topology signal with size or pixel-overlap alone}: only predictions that are correctly typed, spatially overlapping a ground-truth error, and within the size budget receive topological credit, and only insofar as they actually capture the topologically critical skeleton structure.

\section{Experiments}
\label{sec:exp}
\subsection{Experimental Setup}
\label{subsec:exp_setup}
\myparagraph{Datasets.}
We aggregate data from three public sources covering aerial roads~\cite{MnihThesis}, surface cracks~\cite{liu2019deepcrack}, and retinal vasculature~\cite{li2020ipn}, extract $256\times256$ patches via content-aware sampling, and inject four types of topological anomalies through the pipeline of \cref{subsec:data} with Betti-number verification. Error counts are stratified into four complexity levels for curriculum learning. After quality control, the in-distribution split contains $12{,}900$ SFT samples, $50{,}300$ RL samples, and a $4{,}246$-sample test set, \textbf{disjoint at the image level} to rule out patch-level leakage. We additionally curate $1{,}564$ out-of-distribution samples from HALVS leaf-vein and DRIVE retinal-fundus imagery~\cite{ijcai2024p815,staal2004ridge}, which differ in modality, contrast, and color: HALVS introduces unseen leaf textures and reticulate venation, while DRIVE uses fundus photography rather than the training-set OCTA modality. Bounding boxes are normalized to $[0,1000]$; both training data and benchmark will be publicly released.

\myparagraph{Evaluation Metrics.}
A true positive requires both correct type and $\mathrm{IoU}\!\ge\!\tau$ under type-aware Hungarian matching~\cite{lin2014microsoft,kuhn1955hungarian}; we report Precision, Recall, F1 at $\tau\!\in\!\{0.3,0.5,0.75\}$, COCO-style aF1 over $\tau\!\in\![0.50\!:\!0.05\!:\!0.95]$, and Macro F1@0.5. Full definitions are in Appendix~\ref{app:eval_metrics}.

\myparagraph{Models.}
We instantiate Topo-R1 on four open-source backbones with the standard two-stage SFT$+$GRPO pipeline (full-parameter SFT followed by GRPO from the SFT checkpoint): InternVL-2.5-2B~\cite{chen2024internvl}, Qwen2.5-VL-3B~\cite{bai2025qwen2}, Qwen3-VL-4B, and Qwen3-VL-8B~\cite{qwen3technicalreport}, reporting zero-shot, SFT-only, and Topo-R1 numbers for each to isolate per-stage gains. We additionally benchmark four closed-source VLMs (GPT-4o~\cite{hurst2024gpt}, GPT-5.2~\cite{singh2025openai}, Gemini-2.5-Flash~\cite{comanici2025gemini}, Qwen3.5-Plus~\cite{qwen3.5}) under identical zero-shot evaluations. Full implementation details (optimizers, schedules, batch/group sizes) are in Appendix~\ref{sec:supp_impl}.

\begin{figure*}[t]
    \centering
    \includegraphics[width=\textwidth]{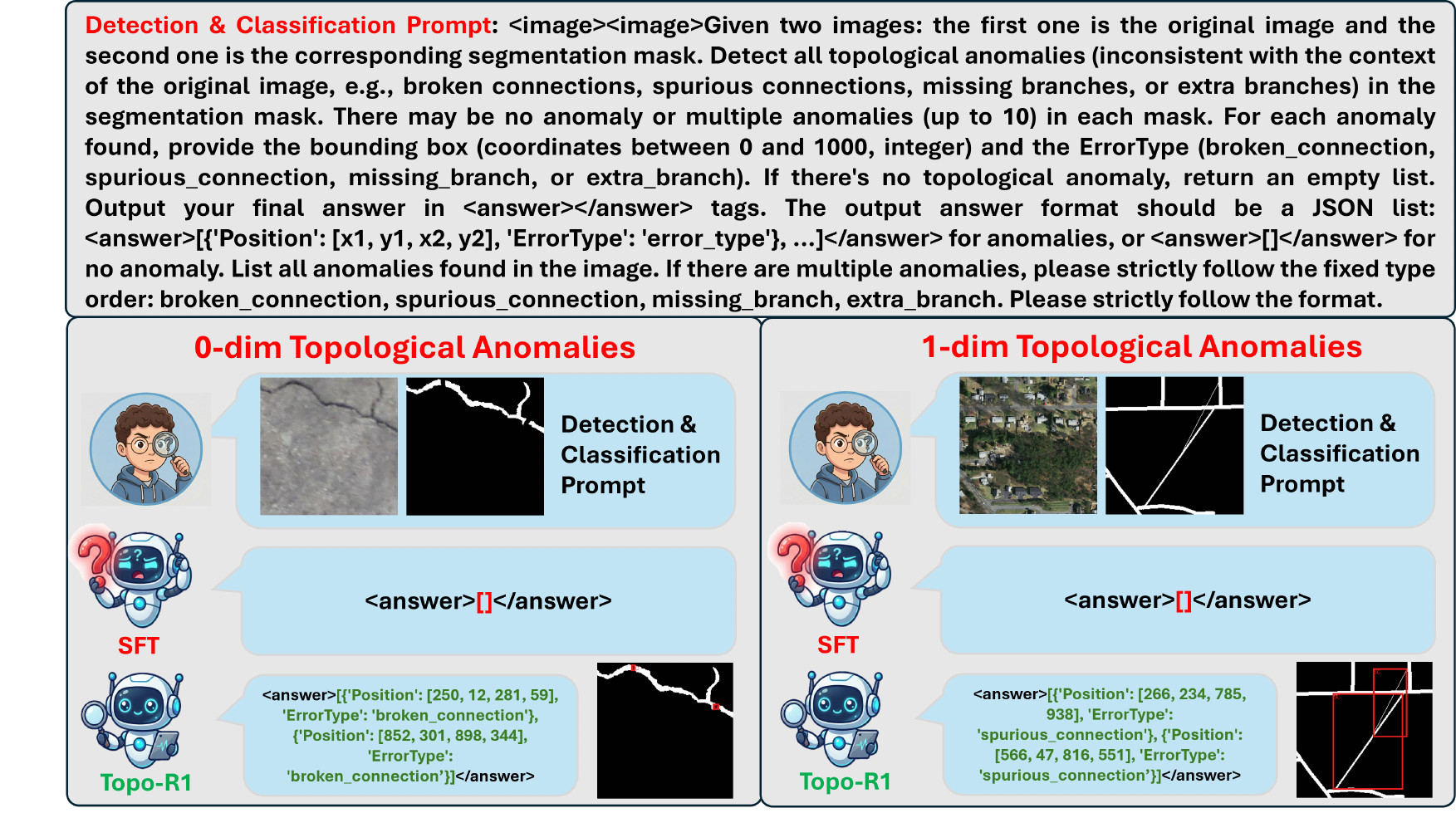}
    \caption{Qualitative results of topological anomaly detection on samples containing 0-dim ($\beta_0$) and 1-dim ($\beta_1$) topological anomalies. \textbf{Topo-R1} demonstrates superior capability in the localization and classification of diverse topological anomalies.}
    \label{fig:qualitative_results}
\end{figure*}

\subsection{Experimental Results}
\label{subsec:results}

\begin{table*}[t]
\centering
\setlength{\tabcolsep}{3.5pt}
\caption{
  Main results on topological anomaly detection on the in-distribution test set.
  We report detection precision (P), recall (R), and F1 at IoU thresholds 0.3, 0.5, and 0.75, along with the COCO-style average F1 (aF1, IoU 0.5:0.95).
  Closed-source models are evaluated in a zero-shot setting.
  ``ZS'' denotes zero-shot inference, ``Sup.'' denotes supervised training of a dataset-specific detector on our anomaly bounding-box labels, ``SFT'' denotes supervised fine-tuning, and ``Topo-R1'' denotes our topology-aware reinforcement fine-tuning.
  Bold indicates the best result in each block.
}
\scriptsize
\begin{tabular*}{\textwidth}{@{\extracolsep{\fill}} lll ccc ccc ccc c}
\toprule
& & & \multicolumn{3}{c}{\textbf{@IoU=0.3}} & \multicolumn{3}{c}{\textbf{@IoU=0.5}} & \multicolumn{3}{c}{\textbf{@IoU=0.75}} & \\
\cmidrule(lr){4-6} \cmidrule(lr){7-9} \cmidrule(lr){10-12}
\textbf{Category} & \textbf{Model} & \textbf{Method} & P\,$\uparrow$ & R\,$\uparrow$ & F1\,$\uparrow$ & P\,$\uparrow$ & R\,$\uparrow$ & F1\,$\uparrow$ & P\,$\uparrow$ & R\,$\uparrow$ & F1\,$\uparrow$ & \textbf{aF1\,$\uparrow$} \\
\midrule
\multirow{4}{*}{\shortstack[l]{Closed-\\Source}}
  & GPT-4o           & ZS & 0.6 & 0.3 & 0.4 & 0.3 & 0.1 & 0.1 & 0.0 & 0.0 & 0.0 & 0.0 \\
  & GPT-5.2          & ZS & 2.4 & 1.7 & 2.0 & 1.0 & 0.7 & 0.8 & 0.2 & 0.1 & 0.1 & 0.3 \\
  & Gemini-2.5-Flash & ZS & 3.0 & 2.4 & 2.7 & 1.0 & 0.9 & 0.9 & 0.1 & 0.1 & 0.1 & 0.3 \\
  & Qwen3.5-Plus     & ZS & 3.1 & 3.2 & 3.2 & 1.4 & 1.5 & 1.5 & 0.2 & 0.2 & 0.2 & 0.5 \\
\midrule
\multirow{3}{*}{\shortstack[l]{Detection /\\VLM}}
  & YOLOv8~\cite{varghese2024yolov8} (img\,$\oplus$\,mask)   & Sup. & 46.5 & 37.5 & 41.5 & 34.5 & 27.6 & 30.7 & 16.5 & 13.1 & 14.6 & 16.4 \\
  & DINO-DETR~\cite{zhang2022dino} (img\,$\oplus$\,mask)     & Sup. & 49.0 & 39.9 & 44.0 & 36.7 & 29.8 & 32.9 & 17.6 & 14.2 & 15.7 & 17.6 \\
  & AnomalyGPT~\cite{gu2024anomalygpt}                       & ZS   &  4.0 &  2.5 &  3.1 &  1.6 &  1.0 &  1.2 &  0.0 &  0.0 &  0.0 &  0.4 \\
\midrule
\multirow{12}{*}{\shortstack[l]{Open-\\Source}}
  & \multirow{3}{*}{InternVL-2.5-2B}
    & ZS       &  0.0 &  0.0 &  0.0 &  0.0 &  0.0 &  0.0 &  0.0 &  0.0 &  0.0 &  0.0 \\
  &  & SFT     & 15.9 & 16.6 & 16.3 &  8.8 &  9.2 &  9.0 &  3.2 &  3.3 &  3.2 &  4.0 \\
  &  & \textbf{Topo-R1} & \textbf{47.3} & \textbf{37.8} & \textbf{42.0} & \textbf{39.2} & \textbf{27.0} & \textbf{32.0} & \textbf{19.2} & \textbf{12.8} & \textbf{15.3} & \textbf{17.1} \\
\cmidrule(lr){2-13}
  & \multirow{3}{*}{Qwen2.5-VL-3B}
    & ZS       &  0.0 &  0.0 &  0.0 &  0.0 &  0.0 &  0.0 &  0.0 &  0.0 &  0.0 &  0.0 \\
  &  & SFT     & 20.9 & 16.4 & 18.4 & 13.6 & 10.7 & 11.9 &  4.7 &  3.7 &  4.1 &  5.3 \\
  &  & \textbf{Topo-R1} & \textbf{66.8} & \textbf{50.9} & \textbf{57.8} & \textbf{49.7} & \textbf{37.9} & \textbf{43.0} & \textbf{21.3} & \textbf{16.2} & \textbf{18.4} & \textbf{21.4} \\
\cmidrule(lr){2-13}
  & \multirow{3}{*}{Qwen3-VL-4B}
    & ZS       &  0.1 &  0.1 &  0.1 &  0.0 &  0.0 &  0.0 &  0.0 &  0.0 &  0.0 &  0.0 \\
  &  & SFT     & 42.6 & 25.5 & 31.9 & 30.7 & 18.3 & 23.0 & 16.2 &  9.7 & 12.1 & 12.8 \\
  &  & \textbf{Topo-R1} & \textbf{70.0} & \textbf{49.9} & \textbf{58.3} & \textbf{54.3} & \textbf{38.7} & \textbf{45.2} & \textbf{27.0} & \textbf{19.3} & \textbf{22.5} & \textbf{24.7} \\
\cmidrule(lr){2-13}
  & \multirow{3}{*}{Qwen3-VL-8B}
    & ZS       &  0.0 &  0.0 &  0.0 &  0.0 &  0.0 &  0.0 &  0.0 &  0.0 &  0.0 &  0.0 \\
  &  & SFT     & 38.4 & 20.5 & 26.7 & 29.0 & 15.5 & 20.2 & 16.9 &  9.0 & 11.8 & 11.8 \\
  &  & \textbf{Topo-R1} & \textbf{69.1} & \textbf{48.5} & \textbf{57.0} & \textbf{53.1} & \textbf{37.3} & \textbf{43.9} & \textbf{27.1} & \textbf{19.0} & \textbf{22.4} & \textbf{24.3} \\
\bottomrule
\end{tabular*}
\label{tab:main_results}
\end{table*}

\myparagraph{Qualitative Results.} As shown in~\cref{fig:qualitative_results}, Topo-R1 localizes and classifies anomalies across both Betti dimensions ($\beta_0$-level disconnections and bridges, $\beta_1$-level spurious loops). The SFT baseline tends to default to empty predictions on these hard cases, an artifact of next-token likelihood training: with no explicit reward for partial recovery, predicting an empty list is the loss-minimizing choice under uncertainty, especially given the $20\%$ negative-sample stratum. Topology-aware RL is therefore essential to bridge the perception gap. A more extensive qualitative gallery and a per-type failure-mode analysis are in Appendix~\ref{sec:supp_failure}.

\myparagraph{Main Quantitative Results.}
\cref{tab:main_results} reveals three patterns: (i) zero-shot performance is uniformly negligible (even the strongest closed-source model reaches only $1.5\%$ F1@0.5), and 1-/3-/5-shot in-context learning likewise fails to elicit topology-aware perception, with the best few-shot result at only $0.5\%$ F1@0.5 (\cref{tab:fewshot}, Appendix~\ref{sec:addi_exps}); (ii) SFT provides a necessary foundation by teaching the anomaly taxonomy and basic localization; and (iii) RL with our composite reward yields consistent gains over SFT across all backbones, particularly in precision. Even the smallest \textbf{Topo-R1} backbone surpasses all closed-source models by an order of magnitude, confirming that topology-aware supervision---rather than model scale or in-context prompting---is the decisive factor.

\myparagraph{Per-Anomaly-Type Breakdown.}
Across all four anomaly types (\cref{tab:per_type}), Topo-R1 substantially outperforms SFT, with the largest gains on broken connections and extra branches---the harder categories with subtler pixel-level cues---indicating that the topology-aware reward steers the model toward fine-grained structural reasoning rather than only the easiest visual cues, particularly for broken connections, where the gap is often only a handful of pixels and lacks salient color cues that would otherwise dominate detection.

\begin{table*}[t]
\centering
\setlength{\tabcolsep}{3.5pt}
\renewcommand{\arraystretch}{1.0}
\caption{
  Detection F1 (\%) by anomaly type at IoU=0.5. Higher is better.
  BC = Broken Connection, SC = Spurious Connection, MB = Missing Branch, EB = Extra Branch.
  Macro is the unweighted average across the four types.
}
\label{tab:per_type}
\scriptsize
\begin{tabular*}{\textwidth}{@{\extracolsep{\fill}} lll ccccc}
\toprule
\textbf{Category} & \textbf{Model} & \textbf{Method} & \textbf{BC} & \textbf{SC} & \textbf{MB} & \textbf{EB} & \textbf{Macro\,$\uparrow$} \\
\midrule
\multirow{4}{*}{\shortstack[l]{Closed-\\Source}}
  & GPT-4o           & ZS & 0.0 & 0.6 & 0.0 & 0.1 & 0.2 \\
  & GPT-5.2          & ZS & 0.0 & 3.5 & 0.0 & 0.1 & 0.9 \\
  & Gemini-2.5-Flash & ZS & 0.1 & 4.2 & 0.1 & 0.0 & 1.1 \\
  & Qwen3.5-Plus     & ZS & 0.1 & 7.2 & 0.0 & 0.0 & 1.8 \\
\midrule
\multirow{12}{*}{\shortstack[l]{Open-\\Source}}
  & \multirow{3}{*}{InternVL-2.5-2B}
    & ZS      &  0.0 &  0.0 &  0.0 &  0.0 &  0.0 \\
  &  & SFT     &  1.6 & 24.5 &  4.2 &  1.9 &  8.1 \\
  &  & \textbf{Topo-R1} & \textbf{16.2} & \textbf{46.5} & \textbf{25.8} & \textbf{12.8} & \textbf{25.3} \\
\cmidrule(lr){2-8}
  & \multirow{3}{*}{Qwen2.5-VL-3B}
    & ZS      &  0.0 &  0.0 &  0.0 &  0.0 &  0.0 \\
  &  & SFT     &  0.6 & 36.9 &  1.0 &  0.7 &  9.8 \\
  &  & \textbf{Topo-R1} & \textbf{30.3} & \textbf{61.8} & \textbf{42.2} & \textbf{22.0} & \textbf{39.1} \\
\cmidrule(lr){2-8}
  & \multirow{3}{*}{Qwen3-VL-4B}
    & ZS      &  0.0 &  0.0 &  0.0 &  0.0 &  0.0 \\
  &  & SFT     & 12.2 & 49.1 &  9.6 &  8.1 & 19.7 \\
  &  & \textbf{Topo-R1} & \textbf{35.4} & \textbf{64.8} & \textbf{39.9} & \textbf{25.7} & \textbf{41.4} \\
\cmidrule(lr){2-8}
  & \multirow{3}{*}{Qwen3-VL-8B}
    & ZS      &  0.0 &  0.0 &  0.0 &  0.0 &  0.0 \\
  &  & SFT     &  3.2 & 49.9 &  3.5 &  8.0 & 16.2 \\
  &  & \textbf{Topo-R1} & \textbf{31.9} & \textbf{66.0} & \textbf{36.3} & \textbf{27.9} & \textbf{40.5} \\
\bottomrule
\end{tabular*}
\end{table*}

\begin{table*}[t]
\centering
\setlength{\tabcolsep}{4pt}
\renewcommand{\arraystretch}{0.92}
\caption{
  Out-of-distribution generalization and evaluation on real segmentation-model outputs.
  All methods receive identical $(I, M)$ inputs at test time and have no access to ground-truth masks. \emph{OOD}: held-out HALVS leaf-vein and DRIVE retinal-fundus imagery; \emph{Real-world}: nnU-Net predictions on Roads and Crack. Topo-R1 is instantiated on Qwen3-VL-4B. Numbers in \%; bold: best per block.
}
\label{tab:beyond_vlm_ood_real}
\scriptsize
\begin{tabular*}{\textwidth}{@{\extracolsep{\fill}} l l ccc cc}
\toprule
\textbf{Setting} & \textbf{Method} & \textbf{F1@.3\,$\uparrow$} & \textbf{F1@.5\,$\uparrow$} & \textbf{F1@.75\,$\uparrow$} & \textbf{aF1\,$\uparrow$} & \textbf{Macro F1@.5\,$\uparrow$} \\
\midrule
\multirow{3}{*}{OOD}
  & Zero-shot (Qwen3-VL-4B)        &  0.1 &  0.0 &  0.0 &  0.0 &  0.0 \\
  & SFT only                       & 33.9 & 24.4 & 14.8 & 14.9 & 21.8 \\
  & \textbf{Topo-R1}               & \textbf{58.4} & \textbf{45.5} & \textbf{23.2} & \textbf{25.5} & \textbf{43.2} \\
\midrule
\multirow{4}{*}{\shortstack[l]{Real-\\world}}
  & GPT-5.2 (zero-shot)              &  1.8 &  0.6 &  0.0 &  0.2 &  0.4 \\
  & Gemini-2.5-Flash (zero-shot)     &  2.1 &  0.8 &  0.0 &  0.3 &  0.5 \\
  & SFT only                         & 26.4 & 18.1 &  7.8 &  9.3 & 15.6 \\
  & \textbf{Topo-R1}                 & \textbf{52.7} & \textbf{39.8} & \textbf{18.4} & \textbf{20.9} & \textbf{36.3} \\
\bottomrule
\end{tabular*}
\end{table*}

\myparagraph{Detection / VLM Baselines.}
To verify that gains are not merely a VLM-backbone artefact, we add three GT-free baselines to \cref{tab:main_results} (\emph{Detection / VLM} block): YOLOv8~\cite{varghese2024yolov8} and DINO-DETR~\cite{zhang2022dino} trained on 4-channel inputs with our anomaly labels, and the anomaly VLM AnomalyGPT~\cite{gu2024anomalygpt} run zero-shot (oracle persistent-homology / skeleton-graph pipelines are excluded as they coincide with our label-generation procedure; details in Appendix~\ref{sec:supp_baselines}). Topo-R1 with mid-to-large backbones (Qwen2.5/3-VL) outperforms every baseline by a clear margin; even the smallest Topo-R1 instantiation (InternVL-2.5-2B) matches the strongest supervised detector despite operating without a dedicated detection head. Together, these results indicate that language-grounded reasoning captures topological inconsistencies that detector-only approaches miss.

\myparagraph{Out-of-Distribution Generalization.}
On the $1{,}564$ OOD samples described in \cref{subsec:exp_setup} (HALVS leaf-vein and DRIVE retinal-fundus imagery; full composition in Appendix~\ref{sec:ood_results}), Topo-R1 retains a clear advantage over SFT and the zero-shot VLM (\emph{OOD} block of \cref{tab:beyond_vlm_ood_real}), indicating that the learned topology-aware perception is a transferable concept rather than a memorized dataset bias.

\myparagraph{Stress Test on Real Segmentation-Model Outputs.}
Beyond the released benchmark, we stress-test Topo-R1 on ${\sim}600$ samples built from a pretrained nnU-Net's~\cite{isensee2021nnu} predictions on held-out Roads and Crack images, with reference annotations derived by symmetric topological differencing (procedure in Appendix~\ref{sec:supp_realworld}). As shown in the \emph{Real-world} block of \cref{tab:beyond_vlm_ood_real}, Topo-R1 maintains a substantial margin over closed-source VLMs and SFT, indicating that the synthetic-to-real gap does not collapse the learned signal and the method can flag candidate regions in real segmentation pipelines.

\myparagraph{Cross-Domain and Sample-Level Robustness.}
Beyond the headline metrics, we further analyze Topo-R1 along several orthogonal axes (full tables in Appendix~\ref{sec:addi_exps}). Across the three in-distribution imaging domains (\cref{tab:per_dataset}), Topo-R1 yields consistent gains over SFT, indicating that the topology-aware reward generalizes across structurally diverse modalities rather than overfitting to a single source. At the per-sample level (\cref{tab:sample-level-metrics}), Topo-R1 dominates anomaly-count accuracy and mean per-sample F1, showing the aggregate gains are not driven by a few easy samples. As anomaly density grows to $6$--$10$ errors per sample (\cref{tab:complexity}), Topo-R1 maintains a substantial F1 margin over SFT, peaking at moderate complexity (2--5 errors). A leave-one-domain-out evaluation (\cref{tab:supp_loo}, Appendix~\ref{sec:supp_loo}) further shows Topo-R1 doubles or triples SFT-only F1 on a held-out domain, confirming that the learned notion of topological inconsistency transfers to unseen modalities.

\subsection{Ablation Studies}
\label{subsec:ablation_study}

\begin{table*}[!htbp]
\centering
\scriptsize
\begin{minipage}[t]{0.38\textwidth}
\centering
\setlength{\tabcolsep}{3pt}
\renewcommand{\arraystretch}{0.95}
\caption{Training-regime ablation on Qwen3-VL-4B: SFT-only, RL-only, and the proposed two-stage pipeline.}
\label{tab:sft_rl_ablation}
\begin{tabular*}{\linewidth}{@{\extracolsep{\fill}} l cccc}
\toprule
\textbf{Regime} & \textbf{F1@.3} & \textbf{F1@.5} & \textbf{F1@.75} & \textbf{aF1} \\
\midrule
RL only             &  5.8 &  3.1 &  0.5 &  1.0 \\
SFT only            & 31.9 & 23.0 & 12.1 & 12.8 \\
\textbf{SFT$+$RL (ours)} & \textbf{58.3} & \textbf{45.2} & \textbf{22.5} & \textbf{24.7} \\
\bottomrule
\end{tabular*}
\end{minipage}
\hfill
\begin{minipage}[t]{0.60\textwidth}
\centering
\setlength{\tabcolsep}{3pt}
\renewcommand{\arraystretch}{0.95}
\caption{Reward component ablation on Qwen3-VL-4B. $R_{\mathrm{fmt}}$ always enabled; $R_{\mathrm{acc}}$: accuracy; $R_{\mathrm{topo}}$: topological (clDice). All metrics in \%; bold: best per column.}
\label{tab:reward_ablation}
\begin{tabular*}{\linewidth}{@{\extracolsep{\fill}} ccc ccccc}
\toprule
$R_{\mathrm{fmt}}$ & $R_{\mathrm{acc}}$ & $R_{\mathrm{topo}}$ & \textbf{F1@.5} & \textbf{aF1} & \textbf{Macro F1@.5} & \textbf{F1@.75} & \textbf{mPS-F1@.5} \\
\midrule
\checkmark & -- & -- & 25.1 & 13.4 & 20.8 & 12.6 & 39.2 \\
\checkmark & \checkmark & -- & 43.8 & 23.5 & 39.7 & 20.1 & 56.8 \\
\checkmark & -- & \checkmark & 31.4 & 17.6 & 26.3 & 17.8 & 44.1 \\
\checkmark & \checkmark & \checkmark & \textbf{45.2} & \textbf{24.7} & \textbf{41.4} & \textbf{22.5} & \textbf{58.5} \\
\bottomrule
\end{tabular*}
\end{minipage}
\end{table*}

\myparagraph{Comparison of SFT-Only, RL-Only, and the Two-Stage Pipeline.}
On Qwen3-VL-4B (\cref{tab:sft_rl_ablation}), RL-only barely improves over zero-shot and remains far below SFT-only: starting from near-zero F1, the policy rarely produces schema-compliant outputs, which collapses the format-gated reward across all rollouts and eliminates the within-group variance GRPO needs~\cite{guo2025deepseekr1}. SFT therefore serves as a \emph{necessary cold start}, after which GRPO further lifts performance to yield the full Topo-R1 result.

\myparagraph{Reward Component Ablation.}
\cref{tab:reward_ablation} confirms that both terms contribute: removing $R_{\mathrm{topo}}$ degrades localization (F1@0.75, aF1) while preserving detection F1@0.5; removing $R_{\mathrm{acc}}$ collapses detection and classification; the full reward dominates every metric.

\section{Conclusion}
\label{sec:conclusion}
We introduced \textbf{Topo-R1}, the first vision-language framework for detecting and classifying topological anomalies in tubular structures.
We built an automated pipeline that injects four controlled topological anomalies with topological verification, trained a VLM via SFT, then GRPO, with a composite reward that combines format consistency, decoupled detection accuracy with type-aware Hungarian matching, and a type-conditioned clDice term.
\textbf{Topo-R1} substantially outperforms all baseline VLMs across multiple detection metrics and generalizes across domains, demonstrating that topology-aware reward shaping enables fine-grained structural understanding in vision-language models.

\clearpage  % TODO FINAL: This \clearpage needs to be removed from both review and camera-ready versions.

% \section*{Acknowledgements}
% Please insert your acknowledgments here.

\bibliographystyle{splncs04}
\bibliography{main}

\newpage
\begin{center}
{\LARGE\bfseries Topo-R1: Detecting Topological Anomalies via Vision-Language Models}\\[0.7em]
{\large ---Appendix---}
\end{center}
\bigskip

\setcounter{section}{5}
\setcounter{table}{5}
\setcounter{figure}{4}

\section*{Appendix Overview}
For ease of navigation, this appendix is organized as follows.
\textbf{Appendix~\ref{app:eval_metrics}} formalizes the evaluation protocol, including the type-aware Hungarian matching, detection/classification/sample-level metrics, and localization quality.
\textbf{Appendix~\ref{sec:supp_impl}} provides implementation details that complement the main paper's reward and training description, including the GRPO objective, accuracy and clDice rewards, the closed-form localization penalty, hyperparameter summary, and compute resources.
\textbf{Appendix~\ref{sec:addi_exps}} reports additional experiments: full few-shot tables, sample-level metrics (count accuracy, mPS-F1), per-dataset and per-complexity breakdowns, the SFT$\to$Topo-R1 improvement summary, and ablations on raw IoU vs.\ tiered reward and on threshold selection.
\textbf{Appendix~\ref{sec:supp_split}} describes test-set generation and splits: the image-level held-out split, train/test distribution match, error-count strata, injection operators, topological verification, and bounding-box formatting.
\textbf{Appendix~\ref{sec:supp_baselines}} details the detection and VLM baselines (YOLOv8, DINO-DETR, AnomalyGPT) used in the main results table.
\textbf{Appendix~\ref{sec:supp_sft_fraction}} ablates the SFT data fraction.
\textbf{Appendix~\ref{sec:supp_loops}} addresses robustness to ground-truth loops.
\textbf{Appendix~\ref{sec:supp_loo}} reports cross-domain leave-one-out generalization.
\textbf{Appendix~\ref{sec:ood_results}} provides full out-of-distribution results, including dataset composition, quantitative metrics, and qualitative predictions on HALVS leaf-vein imagery.
\textbf{Appendix~\ref{sec:supp_realworld}} reports the stress test on real nnU-Net segmentation outputs.
\textbf{Appendix~\ref{sec:supp_failure}} discusses limitations, failure modes, and the planned qualitative gallery.
\textbf{Appendix~\ref{sec:supp_impact}} discusses positive and negative broader impacts and mitigation strategies.

\section{Evaluation Metrics and Protocol Details}
\label{app:eval_metrics}
This appendix provides formal definitions of the evaluation protocol and all metrics used in our experiments.
The model must produce a set of anomaly detections, each consisting of a bounding box $b=[x_1,y_1,x_2,y_2]$ (normalized to $[0,1000]\times[0,1000]$) and a type label $t\in\mathcal{T}$, where
\[
  \mathcal{T}=\{\texttt{broken\_conn.},\;\texttt{spurious\_conn.},\;
               \texttt{missing\_br.},\;\texttt{extra\_br.}\}.
\]
Let $\{g_i\}_{i=1}^{N_g}$ and $\{p_j\}_{j=1}^{N_p}$ denote the ground-truth and predicted anomalies for a given sample, respectively.

\myparagraph{Matching Protocol.}\label{app:hungarian_type_aware} All metrics are built upon a type-aware Hungarian matching that pairs predictions with ground-truth annotations.
For each anomaly type $t\in\mathcal{T}$, we partition both sets by type ($G_t$, $P_t$), construct an IoU cost matrix $C^{(t)}\!\in\!\mathbb{R}^{|G_t|\times|P_t|}$ with $C^{(t)}_{ij}=\mathrm{IoU}(g_i, p_j)$, and solve the optimal linear assignment to maximize total IoU.
A match $(g_i, p_j)$ is accepted only if $\mathrm{IoU}(g_i, p_j) \ge \tau$.
Because matching is performed separately within each type, a prediction that correctly localizes an anomaly but assigns the wrong type is counted as both an FP and an FN, ensuring that classification correctness is strictly enforced.

\myparagraph{Detection Metrics.} Our primary metrics evaluate how well the model detects anomalies at varying localization strictness.
Micro-averaged precision (P), recall (R), and F1 are computed by summing TP, FP, and FN across all test samples:
\begin{equation}
  P = \frac{\mathrm{TP}}{\mathrm{TP}+\mathrm{FP}},\quad
  R = \frac{\mathrm{TP}}{\mathrm{TP}+\mathrm{FN}},\quad
  F1 = \frac{2PR}{P+R}.
\end{equation}
We report P, R, and F1 at IoU thresholds $\tau\in\{0.3, 0.5, 0.75\}$ (\cref{tab:main_results,tab:fewshot}).
To summarize performance across thresholds, we compute the COCO-style average F1: $\text{aF1} = \frac{1}{10}\sum_{k=0}^{9} F1_{\,\tau=0.50+0.05k}$,
which penalizes models that perform well at lenient thresholds but degrade at stricter ones (\cref{tab:main_results,tab:fewshot,tab:ablation_continuous,tab:ablation_threshold}).

\myparagraph{Classification Metrics.} To assess type discrimination beyond aggregate detection, we compute per-type F1 by accumulating TP/FP/FN separately for each anomaly type across all samples. Macro F1 is the unweighted average:
\begin{equation}
  \text{Macro-F1} = \frac{1}{|\mathcal{T}^+|}
    \sum_{t\in\mathcal{T}^+} F1_t,
  \qquad
  \mathcal{T}^+ = \{t : \mathrm{TP}_t + \mathrm{FN}_t > 0\}.
\end{equation}
This ensures that rare types (\eg, \texttt{extra\_branch}, 1181 instances) are weighted equally with common types (\eg, \texttt{broken\_connection}, 4508 instances).
Per-type and Macro F1 are reported in \cref{tab:per_type,tab:improvement,tab:reward_ablation}.

\myparagraph{Sample-Level Metrics.} While the above metrics aggregate across the full test set, we also report per-sample statistics (\cref{tab:sample-level-metrics,tab:ablation_continuous,tab:ablation_threshold}).
\emph{Anomaly Count Accuracy} measures the fraction of samples where the predicted count exactly matches the ground truth: $\frac{1}{N}\sum_{i=1}^{N}\mathbb{1}[N_{\mathrm{pred}}^{(i)}=N_{\mathrm{gt}}^{(i)}]$.
\emph{Anomaly Count MAE} captures the average deviation: $\mathrm{MAE}=\frac{1}{N}\sum_{i=1}^{N}|N_{\mathrm{gt}}^{(i)}-N_{\mathrm{pred}}^{(i)}|$.
\emph{Mean Per-Sample F1 (mPS-F1)} computes F1 per sample using type-aware matching at threshold $\tau$, then averages: $\text{mPS-F1}@\tau = \frac{1}{N}\sum_{i=1}^{N} F1^{(i)}_\tau$.
Note that for anomaly-free samples ($N_{\mathrm{gt}}=N_{\mathrm{pred}}=0$), $F1$ is defined as $1.0$; since 20.4\% of test samples have no anomalies, mPS-F1 is systematically higher than micro-averaged F1.
\emph{Negative Sample Accuracy} measures the fraction of anomaly-free samples where the model correctly predicts $N_{\mathrm{pred}}=0$, reflecting the ability to suppress false alarms.

\myparagraph{Localization Quality.} To measure the spatial precision of correctly detected anomalies, we report the mean IoU among all matched pairs at $\tau=0.5$:
\begin{equation}
  \overline{\mathrm{IoU}} = \frac{1}{|\mathcal{M}|}
    \sum_{(g,p)\in\mathcal{M}} \mathrm{IoU}(g,p),
\end{equation}
where $\mathcal{M}$ is the set of accepted matches. This complements F1 by isolating localization accuracy from detection completeness.
Different anomaly types exhibit different spatial extents: broken connections are typically small ($\sim$10--30\,px), while spurious connections may span $\sim$100--500\,px. For small ground-truth boxes, even minor localization offsets cause IoU to fall below the threshold, partly explaining the consistently higher per-type F1 for spurious connections. We mitigate this by reporting at multiple IoU thresholds, including the lenient $\tau=0.3$.

\section{Implementation Details}
\label{sec:supp_impl}

This section provides additional implementation details that complement the GRPO formulation and the reward formulation described in the main paper (Sec.~\ref{sec:method}).

\myparagraph{GRPO Objective.}
We expand the GRPO update rule summarized in the main paper. Given the per-sample advantages $\hat{A}_i = (R_i - \mu_G)/(\sigma_G + \epsilon_{\mathrm{std}})$ defined in Sec.~\ref{subsec:preliminaries} of the main paper, the policy is updated by maximizing
\begin{equation}
    \mathcal{J}_{\mathrm{GRPO}}(\theta) = \mathbb{E}_{q \sim \mathcal{D},\,\{o_i\}\sim \pi_{\theta_{\mathrm{old}}}(\cdot \mid q)} \left[\frac{1}{G}\sum_{i=1}^{G} L_i^{\mathrm{clip}} \;-\; \beta \, D_{\mathrm{KL}}\!\left(\pi_\theta \,\|\, \pi_{\mathrm{ref}}\right)\right],
\end{equation}
with the per-sample clipped surrogate defined as
\begin{equation}
    L_i^{\mathrm{clip}} = \min\!\Big( r_i(\theta)\,\hat{A}_i,\;\mathrm{clip}\big(r_i(\theta),\,1{-}\varepsilon,\,1{+}\varepsilon\big)\,\hat{A}_i \Big)\,,
\end{equation}
where $r_i(\theta) = \pi_\theta(o_i \mid q) / \pi_{\theta_{\mathrm{old}}}(o_i \mid q)$ is the importance-sampling ratio, with $\theta_{\mathrm{old}}$ denoting the policy parameters from the previous iteration. The clipping threshold $\varepsilon$ bounds the policy update to prevent excessively large steps, while $\beta$ controls the strength of the KL regularization toward the reference policy $\pi_{\mathrm{ref}}$, which is initialized from the SFT checkpoint and held fixed throughout RL training.

\myparagraph{Accuracy Reward.}
Three corner cases are handled before Hungarian matching:
(i)~both $N_{\mathrm{gt}} = 0$ and $N_{\mathrm{pred}} = 0$ (correct negative) yields $R_{\mathrm{acc}} = 1.0$;
(ii)~$N_{\mathrm{gt}} = 0$ but $N_{\mathrm{pred}} > 0$ (all false positives) yields $R_{\mathrm{acc}} = 0.0$;
(iii)~$N_{\mathrm{pred}} = 0$ but $N_{\mathrm{gt}} > 0$ (all false negatives) yields $R_{\mathrm{acc}} = 0.0$.
Otherwise, predictions and ground-truth errors are partitioned by type, and for each type $t \in \mathcal{T}$ an IoU cost matrix is constructed and solved via optimal linear assignment.
Matched pairs with $\mathrm{IoU}_{ij} < \tau_m = 0.1$ are discarded.
Because matching is per-type, type accuracy is implicitly encoded in the assignment.

\myparagraph{clDice Reward.}
For each matched pair $(i, j) \in \mathcal{M}$, we apply \emph{separate} crops to the corrupted and ground-truth masks (and their skeletons) using the predicted and ground-truth bounding boxes, respectively.
Since the two boxes generally differ in size, ground-truth crops are resized to match the predicted-side dimensions via nearest-neighbour interpolation to preserve binary mask values.
The clDice score is then computed on the shape-aligned patches using precomputed skeletons.

\myparagraph{Localization Penalty (Closed Form).}
The closed form of the localization penalty $\mathrm{LocPen}(\cdot)$ summarized in the main paper is:
\begin{equation}
    \mathrm{LocPen}(b) = \begin{cases}
        1 & |b|/|I|\le\tau_{\text{size}}\\[2pt]
        \max\!\Big(0,\; 1-\lambda\,\dfrac{|b|/|I|-\tau_{\text{size}}}{1-\tau_{\text{size}}}\Big) & \text{otherwise,}
    \end{cases}
\end{equation}
with $\tau_{\text{size}}\!=\!0.3$ and $\lambda\!=\!0.8$.

\myparagraph{Hyperparameter Summary.}
Table~\ref{tab:hparams} collects all reward hyperparameters for reproducibility.

\begin{table*}[h]
\centering
\caption{Reward function hyperparameters.}
\label{tab:hparams}
\small
\setlength{\tabcolsep}{8pt}
\begin{tabular*}{\textwidth}{@{\extracolsep{\fill}} lll}
\toprule
\textbf{Parameter} & \textbf{Symbol} & \textbf{Value} \\
\midrule
\multicolumn{3}{l}{\textit{Top-level weights}} \\
Format weight       & $w_{\mathrm{fmt}}$  & 0.10 \\
Accuracy weight     & $w_{\mathrm{acc}}$  & 0.85 \\
Topological weight  & $w_{\mathrm{topo}}$ & 0.05 \\
\midrule
\multicolumn{3}{l}{\textit{Hungarian matching}} \\
Matching IoU threshold & $\tau_m$          & 0.10 \\
\midrule
\multicolumn{3}{l}{\textit{Smooth IoU-to-score mapping $\phi$}} \\
Tier thresholds     & $\{\tau_k\}$        & $\{0.3,\, 0.5,\, 0.7,\, 0.9\}$ \\
Tier reward ceilings & $\{r_k\}$          & $\{0.25,\, 0.55,\, 0.8,\, 1.0\}$ \\
Smoothness exponent    & $\gamma$                 & 1.5 \\
\midrule
\multicolumn{3}{l}{\textit{clDice reward}} \\
Size threshold       & $\tau_{\mathrm{sz}}$ & 0.30 \\
Penalty scale        & $\lambda$            & 0.80 \\
\bottomrule
\end{tabular*}
\end{table*}

\myparagraph{Compute Resources.}
\label{para:supp_compute}
The main experiments are performed on a node of $8\times$NVIDIA RTX A6000 GPUs ($48$\,GB VRAM each), with a small subset of runs (e.g., the largest backbone and selected ablations) executed on a node of $8\times$NVIDIA A100 GPUs ($80$\,GB VRAM each). SFT uses full-parameter fine-tuning across 8 GPUs; GRPO uses 8 GPUs with a group size of $G\!=\!4$ rollouts per query. Closed-source VLMs are evaluated via API.

\section{Implementation Reference}
We acknowledge the open-source codebases used in our implementation.
For supervised fine-tuning, we build upon Qwen2-VL-Finetune~\cite{Qwen2-VL-Finetuning}.
For GRPO training, we use EasyR1~\cite{zheng2025easyr1}, which is built on HybridFlow~\cite{sheng2024hybridflow}.

\section{Additional Experiments}
\label{sec:addi_exps}

We present additional experimental analyses that complement the main results in the paper. These experiments provide fine-grained breakdowns across few-shot prompting, error types, imaging domains, training stages, and sample complexity, offering deeper insight into the strengths and limitations of our approach.

\myparagraph{Few-Shot Evaluation Details.}
\cref{tab:fewshot} reports the full few-shot evaluation across three open-source backbones (Qwen2.5-VL-3B, Qwen3-VL-4B, Qwen3-VL-8B) under zero-, one-, three-, and five-shot settings, summarized in the main paper (\cref{subsec:results}). Even with carefully designed prompts and in-context examples drawn from our training pool, the best result reaches only $0.5\%$ F1@$0.5$, and the precision/recall numbers stay almost indistinguishable from random across IoU thresholds. This confirms that in-context learning alone cannot bridge the gap between general-purpose VLM capabilities and the fine-grained, structured perception required by topological anomaly detection.

\begin{table*}[t]
\centering
\setlength{\tabcolsep}{4pt}
\renewcommand{\arraystretch}{0.88}
\caption{
  \textbf{Few-shot evaluation (1, 3, and 5 in-context examples).}
  We report detection precision (P), recall (R), and F1 at IoU thresholds 0.3, 0.5, and 0.75,
  along with the COCO-style average F1 (aF1, IoU 0.5:0.95). ``ZS'' denotes zero-shot inference. Both zero-shot and few-shot results are almost random, confirming that in-context learning alone cannot elicit topology-aware perception.
}
\label{tab:fewshot}
\scriptsize
\begin{tabular*}{\textwidth}{@{\extracolsep{\fill}} ll ccc ccc ccc c}
\toprule
& & \multicolumn{3}{c}{\textbf{@IoU=0.3}} & \multicolumn{3}{c}{\textbf{@IoU=0.5}} & \multicolumn{3}{c}{\textbf{@IoU=0.75}} & \\
\cmidrule(lr){3-5} \cmidrule(lr){6-8} \cmidrule(lr){9-11}
\textbf{Model} & \textbf{Setting} & P & R & F1 & P & R & F1 & P & R & F1 & \textbf{aF1} \\
\midrule
\multirow{4}{*}{Qwen2.5-VL-3B}
  & ZS     & 0.0 & 0.0 & 0.0 & 0.0 & 0.0 & 0.0 & 0.0 & 0.0 & 0.0 & 0.0 \\
  & 1-shot & 0.7 & 0.1 & 0.1 & 0.3 & 0.0 & 0.0 & 0.0 & 0.0 & 0.0 & 0.0 \\
  & 3-shot & 0.5 & 0.0 & 0.0 & 0.0 & 0.0 & 0.0 & 0.0 & 0.0 & 0.0 & 0.0 \\
  & 5-shot & 2.3 & 0.2 & 0.3 & 0.4 & 0.0 & 0.1 & 0.0 & 0.0 & 0.0 & 0.0 \\
\midrule
\multirow{4}{*}{Qwen3-VL-4B}
  & ZS     & 0.1 & 0.1 & 0.1 & 0.0 & 0.0 & 0.0 & 0.0 & 0.0 & 0.0 & 0.0 \\
  & 1-shot & 1.4 & 1.2 & 1.3 & 0.5 & 0.4 & 0.5 & 0.1 & 0.0 & 0.0 & 0.1 \\
  & 3-shot & 0.7 & 0.7 & 0.7 & 0.2 & 0.1 & 0.2 & 0.0 & 0.0 & 0.0 & 0.0 \\
  & 5-shot & 1.0 & 0.9 & 0.9 & 0.3 & 0.3 & 0.3 & 0.0 & 0.0 & 0.0 & 0.1 \\
\midrule
\multirow{4}{*}{Qwen3-VL-8B}
  & ZS     & 0.0 & 0.0 & 0.0 & 0.0 & 0.0 & 0.0 & 0.0 & 0.0 & 0.0 & 0.0 \\
  & 1-shot & 1.0 & 0.2 & 0.3 & 0.4 & 0.1 & 0.1 & 0.0 & 0.0 & 0.0 & 0.0 \\
  & 3-shot & 0.2 & 0.2 & 0.2 & 0.1 & 0.1 & 0.1 & 0.0 & 0.0 & 0.0 & 0.0 \\
  & 5-shot & 0.9 & 0.5 & 0.6 & 0.3 & 0.2 & 0.2 & 0.0 & 0.0 & 0.0 & 0.1 \\
\bottomrule
\end{tabular*}
\end{table*}

\myparagraph{Sample-Level Metrics.}
Beyond the micro-averaged detection metrics reported in the main paper, \cref{tab:sample-level-metrics} reports per-sample statistics: \emph{Anomaly Count Accuracy} (fraction of test samples whose predicted anomaly count exactly matches the ground-truth count), \emph{Anomaly Count MAE} (mean absolute deviation in count), and \emph{mPS-F1@0.5} (mean per-sample F1 under type-aware matching at IoU$\geq$0.5, which credits partial detection on each sample). Topo-R1 dominates both SFT and the closed-source VLMs across all three sample-level metrics on every backbone, indicating that gains in micro-averaged F1 translate directly into more reliable per-sample outputs rather than being driven by a few easy samples.

\begin{table*}[t]
\centering
\setlength{\tabcolsep}{5pt}
\renewcommand{\arraystretch}{0.88}
\caption{
  Sample-level metrics on the in-distribution test set.
  Count Acc/MAE measures error counting ability (\% and mean absolute error).
  mPS-F1@0.5: mean per-sample F1 at IoU$\geq$0.5 (\%), capturing partial credit.
}
\label{tab:sample-level-metrics}
\scriptsize
\begin{tabular*}{\textwidth}{@{\extracolsep{\fill}} lll ccc}
\toprule
\textbf{Category} & \textbf{Model} & \textbf{Method} & \textbf{Count Acc\,$\uparrow$} & \textbf{Count MAE\,$\downarrow$} & \textbf{mPS-F1@0.5\,$\uparrow$} \\
\midrule
\multirow{4}{*}{\shortstack[l]{Closed-\\Source}}
  & GPT-4o           & ZS & 26.3 & 2.28 & 10.3 \\
  & GPT-5.2          & ZS & 20.3 & 2.24 &  8.6 \\
  & Gemini-2.5-Flash & ZS & 22.0 & 2.35 & 10.5 \\
  & Qwen3.5-Plus     & ZS & 15.5 & 2.27 &  5.2 \\
\midrule
\multirow{12}{*}{\shortstack[l]{Open-\\Source}}
  & \multirow{3}{*}{InternVL-2.5-2B}
    & ZS      & 21.0 & 3.10 & 20.4 \\
  &  & SFT     & 42.8 & 1.65 & 21.9 \\
  &  & \textbf{Topo-R1} & \textbf{46.8} & \textbf{1.28} & \textbf{44.2} \\
\cmidrule(lr){2-6}
  & \multirow{3}{*}{Qwen2.5-VL-3B}
    & ZS      & 13.7 & 3.23 & 13.4 \\
  &  & SFT     & 50.2 & 1.09 & 27.3 \\
  &  & \textbf{Topo-R1} & \textbf{51.5} & \textbf{1.04} & \textbf{56.2} \\
\cmidrule(lr){2-6}
  & \multirow{3}{*}{Qwen3-VL-4B}
    & ZS      & 17.2 & 2.49 &  0.9 \\
  &  & SFT     & 42.0 & 1.55 & 37.7 \\
  &  & \textbf{Topo-R1} & \textbf{54.2} & \textbf{1.02} & \textbf{58.5} \\
\cmidrule(lr){2-6}
  & \multirow{3}{*}{Qwen3-VL-8B}
    & ZS      & 20.4 & 3.18 & 20.4 \\
  &  & SFT     & 41.5 & 1.64 & 35.7 \\
  &  & \textbf{Topo-R1} & \textbf{54.9} & \textbf{1.06} & \textbf{57.5} \\
\bottomrule
\end{tabular*}
\end{table*}

\myparagraph{Per-Dataset Analysis.}
\cref{tab:per_dataset} breaks down detection performance across the four imaging domains in our benchmark. Performance varies across domains: road networks yield the highest F1, likely due to their higher contrast and simpler background, while retinal vasculature (OCTA-3M, OCTA-6M) and crack patterns present greater challenges owing to lower contrast and more complex morphology. Importantly, Topo-R1 delivers consistent improvements over SFT in every domain, confirming that the topology-aware reward generalizes across structurally diverse imaging modalities.

\myparagraph{Improvement Analysis.}
\cref{tab:improvement} quantifies the absolute improvement from SFT to Topo-R1 across three aggregate metrics. The reinforcement learning stage yields large and consistent gains for all backbone models, with the largest single improvement of $+31.1$ overall F1 observed for Qwen2.5-VL-3B. These results confirm that SFT alone provides only a limited foundation, and that GRPO with our topology-aware composite reward is essential for achieving strong performance.

\myparagraph{Effect of Anomaly Complexity.}
\cref{tab:complexity} examines how detection performance scales with the number of ground-truth anomalies per sample. As expected, F1 decreases as the number of anomalies increases from single-anomaly samples to complex multi-anomaly scenes (6--10 anomalies), reflecting the increased difficulty of jointly localizing and classifying multiple anomalies. Nevertheless, Topo-R1 maintains substantially higher performance than SFT across all complexity levels, demonstrating robustness to varying anomaly density.

% Table 5: Per-Dataset Breakdown (F1 @ IoU=0.5)
\begin{table*}[t]
\centering
\setlength{\tabcolsep}{3.5pt}
\renewcommand{\arraystretch}{1.0}
\caption{
  \textbf{Detection F1 (\%) by dataset at IoU=0.5.}
  The benchmark spans four imaging domains:
  road networks (Roads), retinal vasculature at two scales (OCTA-3M, OCTA-6M), and crack patterns (Crack).
}
\label{tab:per_dataset}
\scriptsize
\begin{tabular*}{\textwidth}{@{\extracolsep{\fill}} lll cccc}
\toprule
\textbf{Category} & \textbf{Model} & \textbf{Method} & \textbf{Roads\,$\uparrow$} & \textbf{OCTA-3M\,$\uparrow$} & \textbf{OCTA-6M\,$\uparrow$} & \textbf{Crack\,$\uparrow$} \\
\midrule
\multirow{4}{*}{\shortstack[l]{Closed-\\Source}}
  & GPT-4o           & ZS & 0.1 & 0.1 & 0.3 & 0.1 \\
  & GPT-5.2          & ZS & 0.5 & 1.3 & 1.4 & 0.3 \\
  & Gemini-2.5-Flash & ZS & 0.5 & 2.3 & 1.2 & 0.5 \\
  & Qwen3.5-Plus     & ZS & 0.7 & 3.4 & 2.0 & 0.8 \\
\midrule
\multirow{12}{*}{\shortstack[l]{Open-\\Source}}
  & \multirow{3}{*}{InternVL-2.5-2B}
    & ZS      &  0.0 &  0.0 &  0.0 &  0.0 \\
  &  & SFT     & 11.9 & 12.2 & 10.1 &  7.3 \\
  &  & \textbf{Topo-R1} & \textbf{42.6} & \textbf{28.8} & \textbf{25.3} & \textbf{25.0} \\
\cmidrule(lr){2-7}
  & \multirow{3}{*}{Qwen2.5-VL-3B}
    & ZS      &  0.0 &  0.0 &  0.0 &  0.0 \\
  &  & SFT     & 12.5 & 20.8 & 17.9 &  7.6 \\
  &  & \textbf{Topo-R1} & \textbf{61.4} & \textbf{44.2} & \textbf{39.6} & \textbf{40.1} \\
\cmidrule(lr){2-7}
  & \multirow{3}{*}{Qwen3-VL-4B}
    & ZS      &  0.0 &  0.0 &  0.0 &  0.0 \\
  &  & SFT     & 34.5 & 30.6 & 26.8 & 17.4 \\
  &  & \textbf{Topo-R1} & \textbf{63.8} & \textbf{43.9} & \textbf{40.0} & \textbf{43.5} \\
\cmidrule(lr){2-7}
  & \multirow{3}{*}{Qwen3-VL-8B}
    & ZS      &  0.0 &  0.0 &  0.0 &  0.0 \\
  &  & SFT     & 23.7 & 29.5 & 26.9 & 14.0 \\
  &  & \textbf{Topo-R1} & \textbf{63.9} & \textbf{43.4} & \textbf{39.8} & \textbf{41.1} \\
\bottomrule
\end{tabular*}
\end{table*}

% Table 6: Improvement Analysis (SFT → Topo-R1)
\begin{table*}[t]
\centering
\setlength{\tabcolsep}{3.5pt}
\renewcommand{\arraystretch}{1.0}
\caption{
  \textbf{Absolute improvement ($\Delta$) from SFT to Topo-R1.}
  F1 (\%) at IoU=0.5 for overall, macro (unweighted per-type average), and COCO aF1.
}
\label{tab:improvement}
\scriptsize
\begin{tabular*}{\textwidth}{@{\extracolsep{\fill}} l cc cc cc}
\toprule
& \multicolumn{2}{c}{\textbf{Overall F1@0.5}} & \multicolumn{2}{c}{\textbf{Macro F1@0.5}} & \multicolumn{2}{c}{\textbf{aF1 (COCO)}} \\
\cmidrule(lr){2-3} \cmidrule(lr){4-5} \cmidrule(lr){6-7}
\textbf{Model} & SFT & $\Delta$ & SFT & $\Delta$ & SFT & $\Delta$ \\
\midrule
InternVL-2.5-2B &  9.0 & +23.0          &  8.1 & +17.2          &  4.0 & +13.1 \\
Qwen2.5-VL-3B  & 11.9 & \textbf{+31.1} &  9.8 & \textbf{+29.3} &  5.3 & \textbf{+16.1} \\
Qwen3-VL-4B    & 23.0 & +22.2 & 19.7 & +21.7 & 12.8 & +11.9 \\
Qwen3-VL-8B    & 20.2 & +23.7          & 16.2 & +24.3          & 11.8 & +12.5 \\
\bottomrule
\end{tabular*}
\end{table*}

% Table 7: Performance by Error Complexity (F1 @ IoU=0.5)
\begin{table*}[t]
\centering
\setlength{\tabcolsep}{3.5pt}
\renewcommand{\arraystretch}{1.0}
\caption{
  \textbf{Detection F1 (\%) by anomaly complexity at IoU=0.5.}
  Samples grouped by ground-truth anomaly count: single (n=833), 2--5 (n=1712), 6--10 (n=835).
  Topo-R1 maintains strong performance on complex multi-anomaly samples.
}
\label{tab:complexity}
\scriptsize
\begin{tabular*}{\textwidth}{@{\extracolsep{\fill}} ll ccc}
\toprule
\textbf{Model} & \textbf{Method} & \textbf{Single} & \textbf{2--5} & \textbf{6--10} \\
\midrule
\multirow{2}{*}{InternVL-2.5-2B}
  & SFT     & 18.8 &  9.7 &  7.8 \\
  & \textbf{Topo-R1} & \textbf{35.8} & \textbf{32.4} & \textbf{25.3} \\
\midrule
\multirow{2}{*}{Qwen2.5-VL-3B}
  & SFT     & 13.8 & 12.8 & 11.1 \\
  & \textbf{Topo-R1} & \textbf{47.6} & \textbf{47.1} & \textbf{38.2} \\
\midrule
\multirow{2}{*}{Qwen3-VL-4B}
  & SFT     & 33.2 & 25.5 & 19.4 \\
  & \textbf{Topo-R1} & \textbf{53.7} & \textbf{49.0} & \textbf{40.1} \\
\midrule
\multirow{2}{*}{Qwen3-VL-8B}
  & SFT     & 29.3 & 21.8 & 17.5 \\
  & \textbf{Topo-R1} & \textbf{53.5} & \textbf{47.3} & \textbf{38.7} \\
\bottomrule
\end{tabular*}
\end{table*}

\begin{figure}[h]
  \centering
  \includegraphics[width=0.55\textwidth]{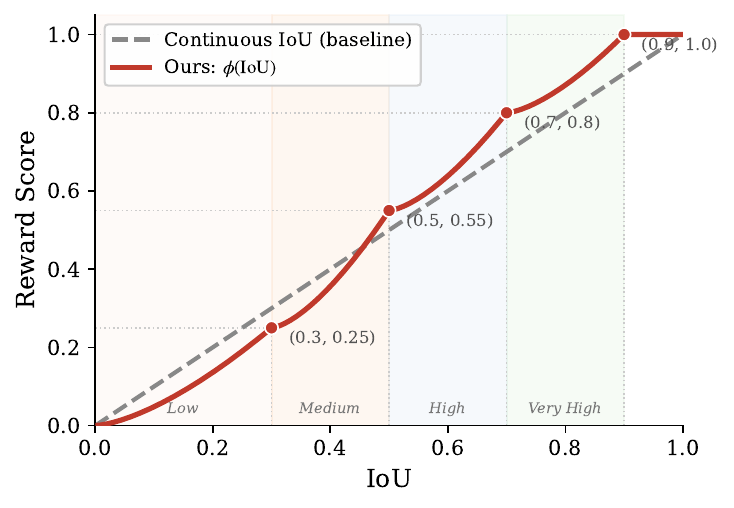}
  \caption{IoU-to-score mapping $\phi$ for different reward designs. Our piecewise non-linear curve (Topo-R1) provides sharp reward shaping near the tier boundaries, while the raw-IoU baseline produces a flat, unshaped slope.}
  \label{fig:reward_mapping}
\end{figure}

\myparagraph{Ablation on Raw IoU Reward.}
We compare our non-linear tiered mapping $\phi$ against a raw-IoU baseline that directly uses the IoU value as the reward score without any piecewise mapping ($\phi(\mathrm{IoU})=\mathrm{IoU}$). The two reward shapes are visualized in \cref{fig:reward_mapping}. As shown in \cref{tab:ablation_continuous}, the raw IoU reward underperforms our design by roughly $27$--$35$ points at F1@0.3 and over $15$ points at F1@0.5 across both backbones, confirming that a flat, unshaped reward signal fails to incentivize the precise localization needed for high-IoU detections.

\begin{table*}[h]
\centering
\renewcommand{\arraystretch}{0.92}
\setlength{\tabcolsep}{4pt}
\caption{
  Ablation on raw IoU vs.\ non-linear tiered reward.
  All metrics in \%; \textbf{bold}: best per backbone.
}
\label{tab:ablation_continuous}
\scriptsize
\begin{tabular*}{\textwidth}{@{\extracolsep{\fill}} ll cccc cc}
\toprule
\textbf{Backbone} & \textbf{Reward} & \textbf{F1@.3} & \textbf{F1@.5} & \textbf{F1@.75} & \textbf{aF1} & \textbf{mPS-F1@.5} & \textbf{Cnt.Acc} \\
\midrule
\multirow{2}{*}{Qwen2.5-VL-3B}
  & Raw IoU            & 22.4 & 14.9 &  5.6 &  6.9 & 29.9 & 50.1 \\
  & \textbf{Topo-R1}   & \textbf{57.8} & \textbf{43.0} & \textbf{18.4} & \textbf{21.4} & \textbf{56.2} & \textbf{51.5} \\
\midrule
\multirow{2}{*}{Qwen3-VL-8B}
  & Raw IoU            & 29.7 & 22.3 & 12.3 & 12.6 & 37.5 & 43.5 \\
  & \textbf{Topo-R1}   & \textbf{57.0} & \textbf{43.9} & \textbf{22.4} & \textbf{24.3} & \textbf{57.5} & \textbf{54.9} \\
\bottomrule
\end{tabular*}
\end{table*}

\myparagraph{Ablation on Threshold Selection.}
We further compare our piecewise non-linear IoU-to-score mapping $\phi$ against two alternative threshold designs: \textbf{Linear}, which uses the same tier boundaries $\{0.3, 0.5, 0.7, 0.9\}$ but with linear interpolation between adjacent tier rewards (no smoothing exponent), and \textbf{COCO}, which uses the standard COCO threshold set $\{0.5, 0.75, 0.9\}$ with linear interpolation. As shown in~\cref{tab:ablation_threshold}, our piecewise non-linear design achieves the best F1 across all IoU levels. Because IoU itself changes non-linearly during training, a linear mapping introduces noisy and inconsistent reward signals, whereas our piecewise non-linear design absorbs this variability within each tier, yielding more stable optimization and stronger overall performance.

\begin{table*}[t]
\centering
\renewcommand{\arraystretch}{0.92}
\setlength{\tabcolsep}{4pt}
\caption{
  Ablation on IoU threshold set (Qwen3-VL-4B).
  \textbf{Linear}: piecewise linear interpolation between tier boundaries; \textbf{COCO}: thresholds $\{0.5,0.75,0.9\}$.
  All metrics in \%; \textbf{bold}: best per column.
}
\label{tab:ablation_threshold}
\scriptsize
\begin{tabular*}{\textwidth}{@{\extracolsep{\fill}} l cccc cc}
\toprule
\textbf{Method} & \textbf{F1@.3} & \textbf{F1@.5} & \textbf{F1@.75} & \textbf{aF1} & \textbf{mPS-F1@.5} & \textbf{Cnt.Acc} \\
\midrule
SFT      & 31.9 & 23.0 & 12.1 & 12.8 & 37.7 & 42.0 \\
Linear   & 56.8 & 43.0 & 20.7 & 23.2 & 57.1 & \textbf{54.9} \\
COCO     & 57.2 & 43.5 & 21.7 & 23.9 & 57.5 & 54.1 \\
\textbf{Topo-R1}     & \textbf{58.3} & \textbf{45.2} & \textbf{22.5} & \textbf{24.7} & \textbf{58.5} & 54.2 \\
\bottomrule
\end{tabular*}
\end{table*}

\section{Test Set Generation and Splits}
\label{sec:supp_split}

\myparagraph{Image-Level Held-Out Split.}
We split each source dataset at the \emph{image} level, not the patch level: every source image is assigned exclusively to either the training pool or the test pool prior to patch extraction. As a consequence, no test patch shares a parent image with any training patch, and the training and test sets cannot overlap by adjacent crops of the same scene. After patch extraction and quality control, the held-out test set contains $4{,}246$ samples drawn from disjoint source images of Roads, Crack, OCTA-3M, and OCTA-6M. The same image-level split convention is used for the leave-one-domain-out experiments and for the OOD/leaf-vein and real-segmentation-output evaluations.

\myparagraph{Train/Test Distribution Match.}
Although the train and test patches are drawn from disjoint source images, both pools follow the same anomaly-injection protocol so that the test distribution is a fair sample of the same task. The anomaly-count strata (zero/single/2--5/6--10) and the per-type sampler are identical between train and test. This is what allows the SFT and SFT$+$RL comparisons to be apples-to-apples; the OOD and real-segmentation-output evaluations test how Topo-R1 behaves when the test distribution shifts \emph{beyond} this protocol.

\myparagraph{Anomaly-Count Strata.}
The number of injected anomalies per sample is drawn from four bins: \emph{0} (${\sim}20\%$, negative samples), \emph{1} (${\sim}20\%$), \emph{2--5} (${\sim}40\%$), and \emph{6--10} (${\sim}20\%$), forming a curriculum of increasing complexity. Within each non-zero bin a balanced sampler preferentially selects underrepresented anomaly types so that the four anomaly classes are represented at roughly equal frequency in the final dataset.

\myparagraph{Injection Operators.}
\emph{Broken connections} erase a short transverse segment of connected skeleton pixels at a randomly sampled interior point; \emph{spurious connections} bridge two distinct components with a thin line; \emph{missing branches} remove pixels within a small region around a skeleton endpoint; and \emph{extra branches} extend a synthetic line from a foreground pixel into the background. The selection of injection sites need not be perfectly precise: the topological verification step (below) automatically rejects any operator application that fails to produce the expected Betti-number change. In multi-anomaly patches, the morphological skeleton is recomputed after each injection to account for prior perturbations.

\myparagraph{Topological Verification.}
For each injection we compute Betti numbers $(\beta_0, \beta_1)$ before and after the modification: a \emph{broken connection} must increase $\beta_0$ or decrease $\beta_1$; a \emph{spurious connection} must decrease $\beta_0$ or increase $\beta_1$; \emph{missing} and \emph{extra branches} must alter either invariant. Failed injections are retried up to $20$ times with a fresh random site; if all retries fail the patch is discarded.

\myparagraph{Bounding-Box Formatting.}
Bounding boxes are derived from the connected components of the pixel-wise difference between the original and corrupted masks, then normalized to $[0,1000]$ and filtered to keep only those whose normalized width and height both lie in $[10,900]$. The ground-truth answer for each sample is a list of dictionaries with \texttt{Position} and \texttt{ErrorType} fields, wrapped in \texttt{<answer>...</answer>} tags and sorted in canonical type order (\emph{broken} $\to$ \emph{spurious} $\to$ \emph{missing} $\to$ \emph{extra}); negative (anomaly-free) samples return an empty list \texttt{[]}.

\section{Detection / VLM Baseline Implementation Details}
\label{sec:supp_baselines}

This section provides implementation details for the detection and VLM baselines reported in Table~\ref{tab:main_results} of the main paper.

\myparagraph{No GT mask at test time.}
All baselines operate in the same setting as Topo-R1: they receive only $(I, M)$ at test time and must decide which regions of $M$ are inconsistent with the structure visible in $I$, with no reference mask and no GT supervision. We deliberately do not include a classical persistent-homology or skeleton-differencing pipeline as a baseline: the oracle form of either method (computing the symmetric topological difference between $M$ and the GT mask) is precisely the procedure used to generate our anomaly labels in Sec.~\ref{subsec:data} and would trivially recover labels at near-$100\%$ F1, while a GT-free heuristic version (e.g., low-persistence + Canny-edge gate) is just a strictly weaker variant of the same operator and provides no independent comparison signal.

\myparagraph{Detector-on-Concat (YOLOv8 / DINO-DETR).}
We train YOLOv8-s and DINO-DETR-R50 on 4-channel inputs constructed by concatenating the RGB image with the binary mask. Both detectors are trained on the same training samples and bounding-box labels as Topo-R1 (single-class detection with a 4-way type head). All hyperparameters follow each detector's default training recipe; we tune learning rate via held-out validation samples drawn from the training pool.

\myparagraph{AnomalyGPT.}
We use the publicly released AnomalyGPT~\cite{gu2024anomalygpt} checkpoint with prompts adapted to ask for typed bounding-box outputs in our schema. No fine-tuning is performed.

\section{SFT Data Fraction Ablation}
\label{sec:supp_sft_fraction}

\myparagraph{Setup.}
To quantify how Topo-R1's gains depend on the amount of supervised fine-tuning data, we vary the SFT subset size $\rho\in\{25\%, 50\%, 100\%\}$ of the $12{,}900$-sample SFT pool, then run the same GRPO stage on the full RL pool. The GRPO data is held fixed across $\rho$. Table~\ref{tab:supp_sft_fraction} reports the results on the in-distribution test set (Qwen3-VL-4B).

\begin{table*}[h]
\centering
\setlength{\tabcolsep}{6pt}
\caption{\textbf{SFT data-fraction ablation} (Qwen3-VL-4B). $\rho$: fraction of SFT samples used; GRPO data fixed.}
\label{tab:supp_sft_fraction}
\small
\begin{tabular*}{\textwidth}{@{\extracolsep{\fill}} l ccc c}
\toprule
\textbf{$\rho$} & \textbf{F1@.3} & \textbf{F1@.5} & \textbf{F1@.75} & \textbf{aF1} \\
\midrule
25\%  & 53.8 & 41.2 & 19.4 & 21.6 \\
50\%  & 56.7 & 43.6 & 21.0 & 23.3 \\
100\% & \textbf{58.3} & \textbf{45.2} & \textbf{22.5} & \textbf{24.7} \\
\bottomrule
\end{tabular*}
\end{table*}

\myparagraph{Observations.}
Topo-R1 remains strong even at $\rho=25\%$, retaining over 91\% of the F1@0.5 of the full-data variant, and degrades only gradually as $\rho$ decreases. This suggests that the SFT stage primarily provides schema bootstrapping; the bulk of the topology-aware capability is shaped by the GRPO stage, in line with the SFT-only / RL-only / two-stage comparison in the main paper.

\section{Robustness to Ground-Truth Loops}
\label{sec:supp_loops}

A natural concern is whether Topo-R1 conflates a \emph{spurious connection} (a bridge that increases $\beta_1$ relative to the clean reference) with a \emph{ground-truth loop} that is part of the true topology, e.g., a closed cycle in a road network or an arteriovenous loop in retinal vasculature.

\myparagraph{Conceptual Distinction.}
By construction of the data pipeline, a spurious connection is always a perturbation \emph{relative to a clean reference mask}: ground-truth loops are never injected and never labeled as spurious. At inference, the model has no access to a reference; it must instead use the original image as the contextual reference and decide whether a bridge in the mask has corresponding visual evidence in the image. Training therefore exposes the model to many clean masks (negative samples and unperturbed regions) that contain legitimate loops, teaching it to associate "loops with corresponding visual support" with the negative class.

\myparagraph{Empirical Test on GT-Loop Samples.}
To test this empirically, we curate a stratified subset of the Roads test pool consisting of samples whose ground-truth mask contains at least one closed cycle (i.e., $\beta_1\ge 1$). We split this subset into two groups and report Topo-R1 (Qwen3-VL-4B) results in Table~\ref{tab:supp_loops}: (i) \emph{clean GT}: GT mask kept as-is, no error injected, measures false-positive rate of Topo-R1 falsely flagging GT loops as spurious; (ii) \emph{injected near GT loop}: a synthetic spurious connection injected in the same patch, measures whether Topo-R1 flags only the injected error and not the GT loop.

\begin{table*}[h]
\centering
\setlength{\tabcolsep}{6pt}
\caption{\textbf{Robustness to ground-truth loops} on Roads (Qwen3-VL-4B). FP-rate on GT loops: fraction of GT loops that Topo-R1 incorrectly flags as spurious; Injected-only F1@.5: F1 measured only against the injected error (lower is better for false alarms on the GT loop, higher is better for detection of the injected error).}
\label{tab:supp_loops}
\small
\begin{tabular*}{\textwidth}{@{\extracolsep{\fill}} l cc}
\toprule
\textbf{Subset} & \textbf{FP-rate on GT loops $\downarrow$} & \textbf{Injected-only F1@.5 $\uparrow$} \\
\midrule
(i) Clean GT (no injection)   & 4.7\% & --- \\
(ii) Injected near GT loop    & 7.3\% & 41.5 \\
\bottomrule
\end{tabular*}
\end{table*}

\myparagraph{Observations.}
Topo-R1 falsely flags only a small fraction of ground-truth loops as spurious (4.7\% on clean GT, 7.3\% when an injected spurious connection is present nearby). When an injection is present, the injected-only F1@0.5 of 41.5 is comparable to the overall F1@0.5 on the full test set (45.2), indicating that the presence of GT loops in the same patch does not collapse Topo-R1's ability to localize and classify the injected error.

\section{Cross-Domain Leave-One-Out Generalization}
\label{sec:supp_loo}

\myparagraph{Setup.}
To probe whether the gains of Topo-R1 transfer to an entirely unseen imaging domain, rather than only benefiting from training on all three domains jointly, we conduct a leave-one-domain-out evaluation. Concretely, we hold out one of $\{$Roads, Crack, Retina$\}$ as the test domain, and train Topo-R1 \emph{only on the remaining two} domains under the standard two-stage pipeline (full-parameter SFT followed by GRPO with our composite reward). This protocol differs from the main per-dataset breakdown (Table~\ref{tab:per_dataset}), in which the model is jointly trained on all three domains. Test data and labels are produced by the same automated pipeline as the in-distribution test set, with image-level held-out splits.

\myparagraph{Results.}
Table~\ref{tab:supp_loo} reports F1@0.5 of SFT-only and Topo-R1 on each held-out test domain (Qwen3-VL-4B). Across all three settings, Topo-R1 consistently outperforms SFT-only by a large margin on the unseen domain, indicating that the topology-aware perception learned from any two source domains transfers to the third without explicit exposure to its imagery during training. The absolute numbers are lower than the in-distribution results in Table~\ref{tab:per_dataset}, as expected for a strictly held-out domain, but the relative gain pattern (Topo-R1 doubles or triples SFT-only) is preserved.

\begin{table*}[h]
\centering
\setlength{\tabcolsep}{6pt}
\caption{\textbf{Leave-one-domain-out generalization (F1@0.5, \%).} The held-out domain is excluded from training; the other two domains are used to train under the standard SFT$+$GRPO pipeline. Backbone: Qwen3-VL-4B.}
\label{tab:supp_loo}
\small
\begin{tabular*}{\textwidth}{@{\extracolsep{\fill}} l ccc}
\toprule
\textbf{Method} & \textbf{Roads (held-out)} & \textbf{Crack (held-out)} & \textbf{Retina (held-out)} \\
\midrule
SFT only           & 21.6 & 11.4 & 16.8 \\
\textbf{Topo-R1}   & \textbf{43.7} & \textbf{30.2} & \textbf{34.6} \\
\bottomrule
\end{tabular*}
\end{table*}

\myparagraph{Discussion.}
The pattern is consistent across the three splits: holding out Crack incurs the largest absolute drop (Crack masks are sparser and less continuous than the other two domains, so two-domain training does not fully cover its statistics), while holding out Roads or Retina yields milder degradation. Importantly, in every case Topo-R1 retains a clear advantage over SFT-only on the unseen domain, which suggests the model's notion of "topological inconsistency between mask and image" is not memorized per dataset but rather learned as a transferable concept.

\section{Out-of-Distribution Results}
\label{sec:ood_results}

\myparagraph{OOD Composition.}
The OOD test set comprises $1{,}564$ $256\!\times\!256$ samples drawn from two sources \textbf{unseen during training} (in-distribution training spans Roads, Crack, OCTA): (i)~\textbf{HALVS}~\cite{ijcai2024p815}, the Hierarchical Annotated Leaf Vein Segmentation dataset, providing $147$ images of Soybean, Sweet Cherry, and London Planetree leaves, sampled into $1{,}364$ samples; and (ii)~\textbf{DRIVE}~\cite{staal2004ridge}, the Digital Retinal Images for Vessel Extraction benchmark, providing $20$ color fundus images (training-split IDs $21$--$40$, held out from any model training in this paper) sampled into $200$ samples. HALVS introduces chlorophyll texture and hierarchical reticulate venation absent from the training pool, and DRIVE differs from the training-set OCTA imagery in modality, contrast, and color statistics, even though both depict retinal vasculature. Source masks are corrupted via the same automated pipeline of Sec.~\ref{subsec:data} with Betti-number verification, stratified into the same four anomaly-count strata as the in-distribution test set; the breakdown is given in Table~\ref{tab:supp_ood_composition}.

\begin{table*}[h]
\centering
\setlength{\tabcolsep}{6pt}
\caption{\textbf{Composition of the OOD test set}: $1{,}564$ samples drawn from HALVS leaf-vein and DRIVE retinal-fundus images, stratified by ground-truth anomaly count per sample.}
\label{tab:supp_ood_composition}
\small
\begin{tabular*}{\textwidth}{@{\extracolsep{\fill}} l ccc l}
\toprule
\textbf{Stratum} & \textbf{\#Samples} & \textbf{HALVS} & \textbf{DRIVE} & \textbf{\#Anomalies / sample} \\
\midrule
\texttt{no\_anomaly}         & 314  & 274 & 40 & 0 \\
\texttt{single\_anomaly}     & 312  & 272 & 40 & 1 \\
\texttt{multi\_anomaly\_2\_5}  & 626  & 546 & 80 & 2--5 \\
\texttt{multi\_anomaly\_6\_10} & 312  & 272 & 40 & 6--10 \\
\midrule
\textbf{Total} & \textbf{1564} & \textbf{1364} & \textbf{200} & --- \\
\bottomrule
\end{tabular*}
\end{table*}

\myparagraph{Quantitative Results.}
Table~\ref{tab:supp_ood} reports detection and per-type metrics on this OOD test set. Topo-R1 retains a large margin over both SFT and the closed-source VLMs, and the per-type pattern (broken / missing / extra harder than spurious) mirrors what is observed on the in-distribution test set, suggesting that the failure modes of Topo-R1 are not domain-specific. Crucially, since the visual statistics of HALVS and DRIVE are fundamentally distinct from our training imagery, this transfer cannot be explained by shared low-level appearance and instead supports the claim that Topo-R1 has learned a transferable notion of topological inconsistency between mask and image.

\begin{table*}[h]
\centering
\setlength{\tabcolsep}{4pt}
\caption{\textbf{OOD evaluation on HALVS leaf-vein and DRIVE retinal-fundus imagery} (Qwen3-VL-4B).}
\label{tab:supp_ood}
\small
\begin{tabular*}{\textwidth}{@{\extracolsep{\fill}} l ccc c c c}
\toprule
\textbf{Method} & \textbf{F1@.3} & \textbf{F1@.5} & \textbf{F1@.75} & \textbf{aF1} & \textbf{Macro F1@.5} & \textbf{mPS-F1@.5} \\
\midrule
GPT-5.2 (zero-shot)         &  2.7 &  1.0 &  0.1 &  0.3 &  0.8 &  6.4 \\
Gemini-2.5-Flash (zero-shot) & 2.9 &  1.2 &  0.1 &  0.4 &  0.9 &  7.1 \\
SFT only                    & 33.9 & 24.4 & 14.8 & 14.9 & 21.8 & 36.6 \\
\textbf{Topo-R1}            & \textbf{58.4} & \textbf{45.5} & \textbf{23.2} & \textbf{25.5} & \textbf{43.2} & \textbf{42.4} \\
\bottomrule
\end{tabular*}
\end{table*}

\begin{figure}[h]
    \centering
    \includegraphics[width=0.85\textwidth]{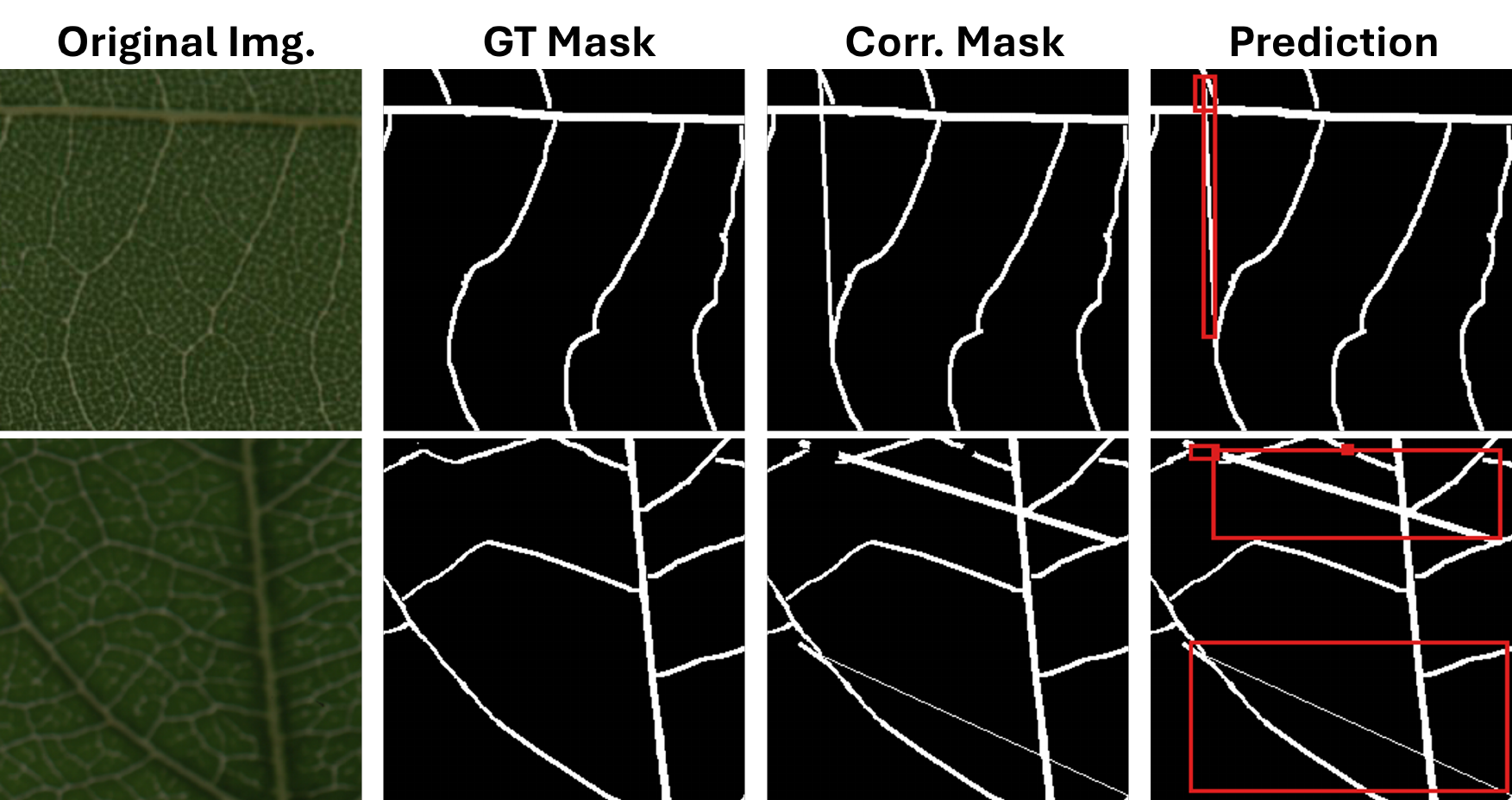}
    \caption{\textbf{Out-of-distribution qualitative results on leaf venation networks.} Topo-R1 generalizes to an unseen domain (HALVS leaf veins) and correctly detects topological anomalies, such as broken and spurious connections, in the corrupted masks. From left to right: original image, ground-truth mask, corrupted mask with injected anomalies, and Topo-R1's prediction. Red boxes denote predicted error regions.}
    \label{fig:ood_qualitative}
\end{figure}

\myparagraph{Qualitative Results on OOD Imagery.}
\cref{fig:ood_qualitative} presents qualitative predictions of Topo-R1 on the held-out HALVS leaf-vein imagery, a domain entirely unseen during training. Despite the marked shift in visual statistics---chlorophyll texture, hierarchical reticulate venation, and natural illumination patterns absent from the training pool---Topo-R1 correctly localizes broken and spurious connections in the corrupted masks, with predicted boxes tightly tracking the topologically-critical skeleton regions rather than dispersing across the leaf surface. This visual evidence corroborates the quantitative gains of Table~\ref{tab:supp_ood}: the learned notion of ``topological inconsistency between mask and image'' transfers from the training modalities (aerial roads, surface cracks, OCTA vasculature) to a structurally analogous but visually distinct domain, supporting the claim that Topo-R1 captures a domain-agnostic structural concept rather than a memorized appearance prior.

\section{Stress Test on Real Segmentation-Model Outputs}
\label{sec:supp_realworld}

This section is \emph{not} part of the released benchmark. It is an additional generalization stress test we run to assess whether Topo-R1's gains transfer from curriculum-style synthetic perturbations to errors actually produced by a trained segmentation model in deployment.

\myparagraph{Setup.}
We run a pretrained nnU-Net~\cite{isensee2021nnu} on held-out Roads and Crack images, producing predicted masks with the statistical character of a deployed segmentation pipeline. For every (image, prediction) pair, we derive reference topological-anomaly annotations by computing the symmetric topological difference between the predicted mask and the ground-truth segmentation, then convert connected residuals to typed bounding boxes following the same protocol as the synthetic in-distribution test set. Predictions whose residual is below a minimum bounding-box size are discarded to remove pixel-level border noise.

\myparagraph{Comparison with Synthetic-Only Errors.}
Compared with synthetic perturbations, the real nnU-Net outputs contain (i) more long, drifting boundary errors that are not localized to a single skeleton site, (ii) many simultaneous small errors that interact, and (iii) type ambiguities (e.g., a thin gap may be a broken connection or a missing branch depending on local context). Topo-R1's gains over SFT are therefore narrower than on the synthetic test set but remain large (\emph{Real-world} block of Table~\ref{tab:beyond_vlm_ood_real} in the main paper), indicating that the model's representation of "topological inconsistency between mask and image" transfers from the curriculum-style synthetic errors to messier real-model failures.

\section{Limitations and Failure Cases}
\label{sec:supp_failure}

\myparagraph{Limitations.}
\label{para:supp_limitations}
Topo-R1 has three main limitations: (i)~it is instantiated only on $2$--$8$B-parameter VLMs, leaving its behavior at much larger scales unverified; (ii)~the input image is the sole contextual reference at test time, so highly ambiguous inputs (e.g., severely occluded vessels) may yield legitimate visual gaps being flagged---we therefore frame outputs as \emph{candidate flags} for downstream verification; and (iii)~training labels come from synthetic injections, and the stress test on real nnU-Net outputs (\cref{sec:supp_realworld}) shows the synthetic-to-real gap is not catastrophic but does narrow the SFT$\to$Topo-R1 margin. We leave scaling, multi-image references, and direct training on real model failures to future work.

\myparagraph{Failure Modes.}
We identify three recurring failure modes of Topo-R1: (i) \emph{type confusion} between extra-branch and spurious-connection when the new edge is short (less than ${\sim}10$ pixels) and aligned with the local skeleton orientation; (ii) \emph{over-segmentation in dense regions}, where multiple injections overlap and Topo-R1 produces a single union bounding-box that covers all of them; (iii) \emph{missed micro-anomalies} smaller than ${\sim}5$ pixels, particularly in OCTA where image contrast at that scale is intrinsically ambiguous. Each of these modes is consistent with the localized-yet-important nature of topological perception discussed in the introduction: the harder cases concentrate at the spatial scale where pixel-level evidence is sparsest.

\myparagraph{Qualitative Gallery.}
A qualitative gallery covering all four anomaly types and all four imaging domains, plus representative failure cases, will be released alongside the benchmark.

\section{Broader Impacts}
\label{sec:supp_impact}

\myparagraph{Positive Impacts.}
Topo-R1 outputs \emph{candidate flags} of potential topological inconsistency in tubular-structure segmentations to assist human expert review. By directing users' attention to suspicious regions, it may accelerate annotation throughput and reduce the per-sample burden on annotators and reviewers in domains where topological correctness is critical, such as biomedical imaging, remote sensing, and infrastructure inspection. The same candidate-flag signal can also feed downstream learning pipelines as a sample-selection prior for active learning and as pseudo-supervision for self-supervised segmentation refinement, while final correction and labeling decisions remain with domain experts.

\myparagraph{Negative Impacts and Mitigations.}
Topo-R1 outputs \emph{candidate flags}, not final judgments; if practitioners treat its outputs as ground-truth corrections without expert review, they may inadvertently introduce errors into safety-critical pipelines (e.g., clinical diagnosis, navigation systems). To mitigate this, we explicitly frame the output as a candidate-flag for downstream verification (Sec.~\ref{subsec:data}) and recommend that any deployment include an expert-in-the-loop verification step. The synthetic-curriculum training also means Topo-R1 may underperform on rare real-world failure modes that fall outside the four anomaly types we synthesize; we encourage downstream users to validate on domain-specific data before deployment.

\end{document}